 \documentclass[11pt,letterpaper]{article}
\usepackage[top=0.75in,bottom=1in,left=0.7in,right=0.7in]{geometry}
\usepackage{amsmath,amsfonts}
\usepackage{array}
\usepackage[caption=false,font=normalsize,labelfont=sf,textfont=sf]{subfig}
\usepackage{xcolor}
\usepackage{bm}
\usepackage{amssymb}

\usepackage{pifont}

\usepackage{balance}
\usepackage{booktabs}
\usepackage{tabularx}
\usepackage{graphicx}
\usepackage{multirow}
\usepackage{algorithm}
\usepackage{url}
\usepackage{algcompatible}

\newtheorem{myTheo}{Theorem}
\newtheorem{myProp}{Proposition}
\newtheorem{assumption}{Assumption}
\newtheorem{lemma}{Lemma}
\usepackage{color}

\usepackage{colortbl}

\author{Jia-Qi~Lin, 
        Yuangang Pan, 
		Chang-Dong~Wang,
		Haizhang~Zhang,\\ Ivor W. Tsang, and Joey Tianyi Zhou}
\date{}
\begin{document}

\title{Reliable Neural Collapse Approximation for Open-World Test-Time Adaptation}



\maketitle

\begin{abstract}
Test-Time Adaptation~(TTA) methods aim to bridge the domain gap between the source and target domains. However, traditional TTA methods become ineffective when the label distribution shift occurs, a challenge commonly referred to as an open-world scenario. In this paper, we introduce a new method named Reliable Neural Collapse approximation~(ReNC) for Open-World Test-Time Adaptation~(OWTTA). 
{Specifically, we leverage neural collapse as a structural prior for reliable target-domain adaptation. Guided by this prior, we justify that the pre-trained classifier weights can serve as the prototypes of the source domain.} By measuring the similarity between samples and prototypes, we filter out the Out-Of-Distribution~(OOD) samples for reliable updates. Furthermore, we propose a neural collapse approximation mechanism to refine these prototypes, ensuring they can gradually adapt to the target domain while maintaining the neural collapse structure. Extensive experiments on several open-world benchmarks demonstrate the superiority of the proposed method. 
{Our empirical analysis suggests that ReNC better preserves NC-related properties in the target domain, providing useful evidence for explaining reliable OWTTA and offering new insights for model design. Code is available at \url{https://github.com/JiaqiLin-AI/ReNC}.}

\footnote{{Jia-Qi Lin, Yuangang Pan, Ivor W. Tsang, and Joey Tianyi Zhou are with the Centre for Frontier AI Research, Agency for Science, Technology and Research (A*STAR), Singapore 138632. E-mails: Lin\_Jiaqi@a-star.edu.sg, Pan\_Yuangang@a-star.edu.sg, ivor\_tsang@a-star.edu.sg, joey\_zhou@a-star.edu.sg. Jia-Qi Lin completed this work during his Ph.D. study at Sun Yat-sen University. Chang-Dong Wang is with School of Computer Science and Engineering, Sun Yat-sen University, Guangzhou, China, Key Laboratory of Machine Intelligence and Advanced Computing, Ministry of Education, China, and Guangdong Province Key Laboratory of Computational Science, Guangzhou, China. E-mail: changdongwang@hotmail.com. Haizhang Zhang is with School of Mathematics (Zhuhai), Sun Yat-sen University, Zhuhai, China. E-mail: zhhaizh2@mail.sysu.edu.cn. Corresponding author: Chang-Dong Wang.}}

\end{abstract}

\section{Introduction}
Test-Time Adaptation (TTA) has attracted significant attention~\cite{DBLP:conf/nips/Ben-DavidBCP06,DBLP:journals/ml/Ben-DavidBCKPV10,DBLP:conf/cvpr/YouLCWJ19} for its ability to improve the generalization of deep learning models without requiring access to the entire batch of data. It aims to bridge the distribution gap between training and testing data under conditions of \textit{data distribution shift}, where the training and testing data originate from different domains. Existing TTA methods~\cite{DBLP:conf/cvpr/BoudiafMAB22, DBLP:conf/icml/NiuW0CZZT22, DBLP:conf/cvpr/ChiWYT21, DBLP:conf/nips/GoyalSRK22} address domain adaptation by either aligning target features with the source distribution or minimizing the entropy of model outputs. Both approaches have shown impressive effectiveness in enhancing model performance under data distribution shifts.

\begin{figure}[t]
\centering   
\subfloat[Existing Methods]{      
\includegraphics[width=0.3\linewidth]{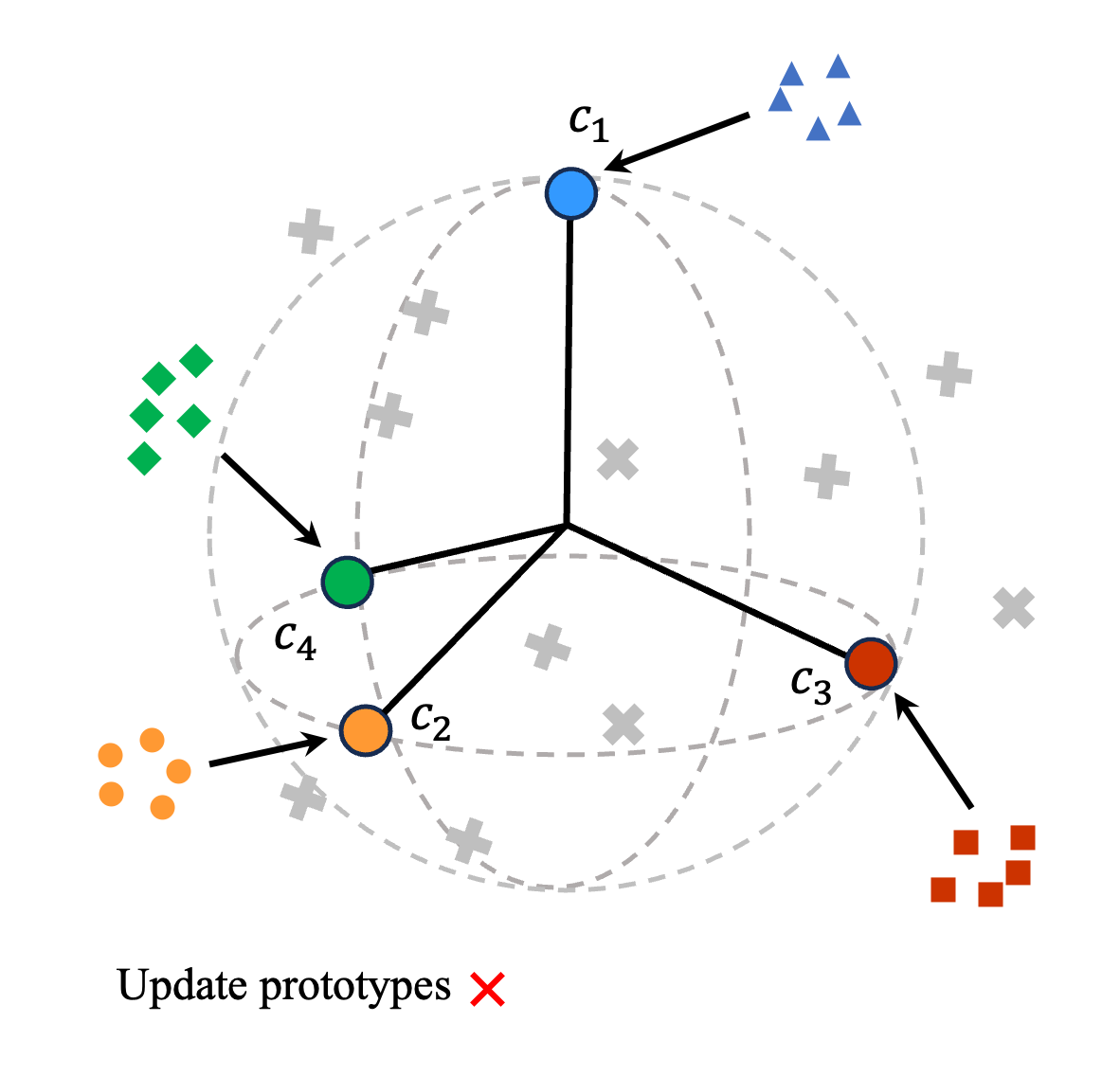} 
\label{fig:motivation_1}
}
\subfloat[ReNC]{      
\includegraphics[width=0.3\linewidth]{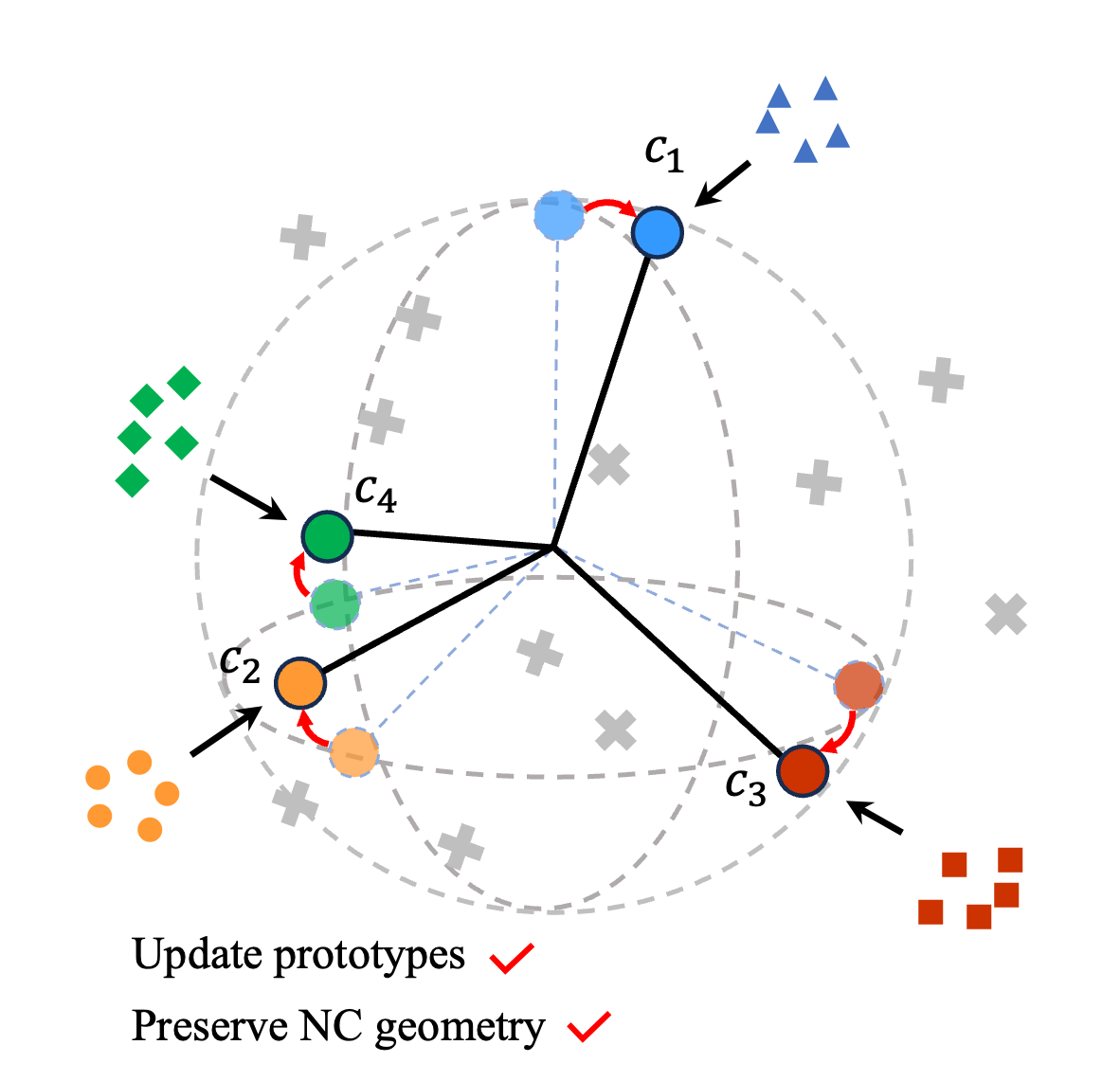}
\label{fig:motivation_2}
}\\
\centering
\subfloat{      
\includegraphics[width=0.4\linewidth]{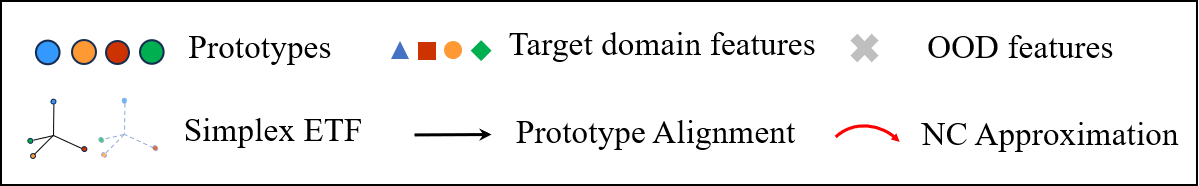}
\label{fig:motivation_3}
}
\caption{{Motivation of ReNC. (a) Existing methods use fixed source-domain prototypes, which may mismatch the target domain. (b) ReNC improves feature-prototype consistency and adapts prototypes toward target-domain neural collapse.}}
\label{fig:motivation}
\end{figure}

A basic assumption of TTA is that the training and testing samples share the same label space.
However, this assumption may be violated in more realistic open-world scenarios~\cite{DBLP:conf/iccv/Li0SJ23}. Under the open-world scenarios, the test data may not only contain seen classes, known as \textit{In-Distribution (ID)} data, but also unseen classes, referred to as \textit{Out-Of-Distribution (OOD)} data. 
This phenomenon, which we refer to as a \textit{label distribution shift}, can render existing TTA methods ineffective.
{For example, in medical imaging, data distribution shift and label distribution shift often occur due to differences in acquisition devices and new disease that the model has not encountered during training~\cite{DBLP:conf/cvpr/YuanXDCYQYYSCL023}.}
Adapting models to simultaneous data distribution shift and label distribution shift is the known as Open-World Test-Time Adaptation~(OWTTA), which remains a challenge for the community. 
There are two primary challenges in OWTTA: 1) The label distribution shift caused by OOD samples renders the model update process unreliable. 2) The domain gap introduced by the data distribution shift further increases the difficulty of effective adaptation.

Most recently, a limited number of studies~\cite{ DBLP:conf/iccv/Li0SJ23, DBLP:conf/iccv/LeeDCC23, gao2024unified} have explored OWTTA as an effective strategy to enhance the generalization abilities of deep neural networks. They typically employ a two-step process~\cite{DBLP:conf/iccv/LeeDCC23, gao2024unified} to address the aforementioned challenges. First, the OOD samples are filtered out from the target data to mitigate the negative effects of label distribution shift. This is typically achieved by measuring the similarity between features and their nearest prototypes to distinguish between ID and OOD samples. Next, TTA is applied to the remaining ID samples to address the distribution gap caused by data distribution shift.
However, these methods optimize the model using fixed prototypes, which are typically derived either from the source domain or the pre-trained classifier. {In practice, source prototypes are often unavailable due to computational constraints~\cite{DBLP:conf/icml/LiangHF20,DBLP:conf/cvpr/KunduVVB20}, privacy concerns~\cite{DBLP:conf/aaai/LiuZH25,DBLP:conf/cvpr/LiJ0WW20}, or copyright restrictions~\cite{DBLP:journals/tip/FengXT21}.} {Moreover, even when accessible, prototypes from the source domain often fail to align well with the target domain due to the data distribution shift.}
As illustrated in Figure~\ref{fig:motivation_1}, 
existing method~\cite{DBLP:conf/iccv/Li0SJ23} push features toward fixed prototypes of the source domain leading to suboptimal performance due to the domain gap with the target domain.

In this paper, we introduce a new OWTTA method, Reliable Neural Collapse Approximation~(ReNC), which addresses the aforementioned challenges by approximating the neural collapse structure in the target domain. {Neural collapse~\cite{DBLP:conf/nips/ZhouY0LL0Z22} describes a terminal geometry of well-trained classification neural networks, where features become intra-class compact, class means align with classifier weights, and classification behaves like nearest-class-center prediction. Motivated by this property, we leverage neural collapse as a reliable structural prior to guide OWTTA. \textit{{However, as neural collapse typically occurs only with labeled source data, it cannot be directly achieved on unlabeled target data.}} Therefore, our ReNC method is proposed to reliably approximate the neural collapse state in the target domain.} 
{As illustrated in Figure~\ref{fig:motivation_2}, ReNC not only pushes target features toward their corresponding prototypes, but also updates the prototypes to approximate the target-domain neural collapse state.}
{Specifically, considering that the pre-trained source model has already collapsed, we introduce a parameter-free mechanism to filter out OOD samples for reliable updates. This mechanism evaluates the similarity of the target domain features to prototypes, ensuring the adaptation process focuses on ID samples.}
{Furthermore, we formulate neural collapse approximation as a progressive adaptation process that encourages reliable target features to approach their corresponding prototypes, while updating the prototypes with prior-regularized adaptation to preserve the underlying neural collapse geometry.}
Benefiting from reliable and efficient updates, ReNC achieves a better neural collapse state in the target domain without access to the entire target data batch. We summarize the contributions as follows:
\begin{itemize}
    \item [$\bullet$] We are the first study to introduce the concept of neural collapse to OWTTA. And we justify that classifier weights are equivalent to source domain prototypes from the neural collapse perspective.


    \item[$\bullet$] We introduce a new OWTTA framework, called ReNC, featuring an efficient updating mechanism to approximate the neural collapse state in the target domain, {which provides a reliable structural prior for TTA.}

    \item[$\bullet$] {Extensive experiments on several open-world benchmarks demonstrate the effectiveness of ReNC in OWTTA, and further analyses show that ReNC better preserves NC-related representation properties during target adaptation.}

\end{itemize}

The rest of this paper is organized as follows: In Section~\ref{sec:problem_def}, we define the problem definition of OWTTA and provide a brief review of the related literature. 
Section~\ref{sec:method} introduces our ReNC method, progressing from the basic concepts to the final approach.
In Section~\ref{sec:experiments}, we present a thorough evaluation based on extensive experiments to validate the effectiveness of the ReNC approach. Finally, Section~\ref{sec:conclusion} provides a summary of the key findings and concludes the paper.

\section{Problem Definition and {Literature} Review}~\label{sec:problem_def}
In this section, we first introduce the setting of closed-world test-time adaptation. Then, we extend the problem to the open-world scenario, where data contains label distribution shifts. Lastly, we compare this setting with OOD detection and universal domain adaptation, discussing the differences between them and OWTTA.

\subsection{Test-Time Adaptation}
\textbf{Data Distribution Shift:} 
We consider the classification tasks where $\mathcal{X}$ and $\mathcal{Y}$ represent the data space and the label space, respectively. There are two distinct distributions over $\mathcal{X} \times \mathcal{Y}$, referred to as the source domain $\mathcal{D}_S$ and the target domain $\mathcal{D}_T$. 
The data distribution shift occurs because the source data set $\mathcal{S}_S$ and the target data set $\mathcal{S}_T$ are drawn independently from two different domains, $\mathcal{D}_S$ and $\mathcal{D}_T$, respectively, resulting in differences in their distributions:
\begin{equation*}
    \mathcal{S}_S=\left\{\left(\mathbf{x}_i,y_i \right) | \mathbf{x}_i \in \mathcal{D}_S \right\}, \quad
    \mathcal{S}_T=\left\{\mathbf{x}_i | \mathbf{x}_i \in \mathcal{D}_T \right\}.
\end{equation*}

Various domain adaptation methods~\cite{10552873,10623315} have been proposed to eliminate the domain gap between $\mathcal{D}_S$ and $\mathcal{D}_T$. Among all these methods, Test-Time Adaptation (TTA)~\cite{DBLP:conf/aaai/WuC0PF24,DBLP:conf/aaai/Su0J24} has gained significant attention for its assumption that the model cannot access source data and target labels, making it more suitable for real-world applications. 

A number of TTA methods are inspired by the self-training and employ various techniques for unlabeled target data adaptation~\cite{DBLP:conf/nips/GoyalSRK22,DBLP:conf/nips/GandelsmanSCE22}. 
Test-time entropy minimization (TENT)~\cite{DBLP:conf/iclr/WangSLOD21} is proposed to update only the parameters in the batch normalization layer by minimizing the entropy of logits.
 Source Hypothesis Transfer~(SHOT)~\cite{DBLP:conf/icml/LiangHF20} freezes the classifier of the source model and updates the feature extraction module by minimizing entropy while maintaining balanced distribution in the target domain.

Besides, the distribution alignment techniques are also utilized in TTA by treating the source domain distribution as an anchor for aligning the features of the target domain~\cite{DBLP:conf/nips/SuXJ22, DBLP:conf/ijcai/KojimaMI22,DBLP:conf/cvpr/MirzaJ0KPB23}. Test-time training method~\cite{DBLP:conf/nips/LiuKDBMA21} modified pre-training procedure to acquire the statistic information of source domain without accessing source data when adaptation. Nevertheless, such a scheme is still impractical since it alternates the pre-training procedure. To tackle this issue, \cite{10452869} improves TTA performance by saving the distribution of the source domain and aligning the high-confidence target features towards the distribution for updating without altering pre-training.

Additionally, self-supervised learning techniques are also widely used in TTA since they do not need to consider the semantic information of the target data~\cite{DBLP:conf/iclr/HansenJSAAEPW21}. For example, contrastive learning is introduced~\cite{DBLP:conf/cvpr/0001WDE22} to refine the online pseudo labeling process, leveraging the self-supervised techniques to enhance the target feature adaptation. The self-supervised training branch is introduced~\cite{DBLP:conf/icml/SunWLMEH20} to fine-tune the parameters of the feature encoder during adaptation.

{While most test-time adaptation (TTA) methods have been developed under the assumption of static target distributions and focus primarily on visual data, recent works have begun exploring more complex and realistic settings. Some studies consider scenarios where the target domain distribution changes continuously over time~\cite{DBLP:conf/cvpr/0013FGD22, DBLP:conf/iclr/Niu00WCZT23,DBLP:conf/nips/GongJKKSL22,DBLP:conf/cvpr/DoblerM023}, and~\cite{DBLP:conf/icml/0007GJZL23} further extend this to evolving environments where the proportions of existing classes vary dynamically, potentially leading to class imbalance. In parallel,~\cite{DBLP:conf/aaai/ZhouYGL25} investigates TTA in the context of tabular data, and proposes a method that estimates the test-time label distribution using confident predictions and calibrates model outputs accordingly. Furthermore, Vision-Language Models (VLMs)~\cite{DBLP:conf/icml/RadfordKHRGASAM21} have also been explored for test-time adaptation through prompt tuning, where prompts are optimized without labels to improve both accuracy and calibration during inference~\cite{DBLP:conf/iclr/YoonYTHLY24}.}

\vspace{-10pt}
\subsection{Open-World Test-Time Adaptation}
\textbf{{Label distribution shift}:}
Existing TTA methods have achieved significant improvement, assuming that the source and target domain have the same label space. {In the open-world scenario, however, this assumption may be violated~\cite{DBLP:conf/icdm/QianBZZZ23}}, with the distribution shift existing in both the data space $\mathcal{X}$ and the label space $\mathcal{Y}$, simultaneously.


 This shift is characterized by the divergence between the source label space and the target label space. 
 The label set of the target domain $\mathcal{Y}_T$ contains not only the labels from the source domain $\mathcal{Y}_S$, but also additional labels from new classes, denoted as $\mathcal{Y}_O$:
\begin{equation*}
    \mathcal{Y}_T=\left\{\mathcal{Y}_S \cup \mathcal{Y}_{O}\right\}, \quad \mathcal{Y}_S \cap \mathcal{Y_O} = \emptyset.
\end{equation*}

Based on the label of samples, the target data set $\mathcal{S}_T$ can be further partitioned into the In-Distribution (ID) sample set $\mathcal{S}_{T_I}$ and the Out-Of-Distribution (OOD) sample set $\mathcal{D}_{T_O}$:
\begin{equation*}
    \begin{aligned}
        \mathcal{S}_{T_I}=&\left\{\mathbf{x}_i | \mathbf{x}_i \in \mathcal{D}_{T_I}, \Psi(\mathbf{x}_i)\in \mathcal{Y}_S \right\},
        \\
    \mathcal{S}_{T_O}=&\left\{\mathbf{x}_i |\mathbf{x}_i \in \mathcal{D}_{T_O}, \Psi(\mathbf{x}_i)\in \mathcal{Y}_O \right\},
    \end{aligned}
\end{equation*}
where $\mathcal{D}_{T_I}$ represents the domain of ID samples, $\mathcal{D}_{T_O}$ represents the domain of OOD samples, and $\Psi$ is the ground truth mapping.
OWTTA requires a robust method to ensure that the adaptation process focuses on the ID samples while minimizing the adverse impact of the OOD samples.

Few TTA methods have taken the open-world scenario into consideration. Open-World Test-Time Training (OWT3\footnote{To avoid confusion, we renamed the OWTTT method to OWT3.})~\cite{DBLP:conf/iccv/Li0SJ23} distinguishes OOD samples based on the distance between target data and prototypes from the source domain. It then expands the prototype pool and aligns the ID samples with statistical information from the source domain for reliable adaptation. 
OSTTA~\cite{DBLP:conf/iccv/LeeDCC23} measures sample confidence based on the probability values of logits. It leverages the wisdom of crowds to select high-confidence samples and applies entropy minimization exclusively to these selected samples.
The Unified Entropy Optimization (UniEnt)~\cite{gao2024unified} method introduces two Gaussian Mixture Models (GMMs) to model ID and OOD samples based on their distance to the nearest prototypes. UniEnt then minimizes the entropy of ID samples while maximizing the entropy of OOD samples to enhance the TTA performance.

\vspace{-10pt}
\subsection{Related Branches}



\textbf{Out-of-Distribution Detection.}
OOD detection aims to identify samples outside the training distribution, which is important for reliable model deployment. Existing methods usually design post-hoc scoring functions based on logits~\cite{DBLP:conf/iclr/HendrycksG17,DBLP:conf/nips/LiuWOL20,DBLP:conf/nips/SunGL21}, feature representations~\cite{DBLP:conf/icml/SunM0L22,DBLP:conf/icml/SastryO20,DBLP:conf/cvpr/Wang0F022}, or gradient information~\cite{DBLP:conf/nips/HuangGL21,DBLP:journals/corr/abs-2205-10439}. Some studies further improve OOD detection by re-training classifiers with auxiliary OOD datasets~\cite{DBLP:conf/cvpr/LiV20,DBLP:conf/icml/MingFL22} or generated synthetic OOD samples~\cite{DBLP:journals/corr/abs-1910-04241,DBLP:conf/iclr/LeeLLS18,DBLP:conf/iclr/DuWCL22}. 
OOD detection methods require access to entire batches of data at once and do not account for data distribution shifts. In contrast, OWTTA must adapt and make inferences instantly with new incoming batches, addressing both data and label distribution shifts simultaneously.

\textbf{Universal Domain Adaptation.} Similar to TTA, existing domain adaptation methods focus on data distribution shifts while neglecting label distribution shifts in the target domain. To provide more flexibility in real-world scenarios, universal (or open-set) domain adaptation methods have been proposed~\cite{DBLP:conf/cvpr/YouLCWJ19,DBLP:conf/nips/ChangSTW22,DBLP:conf/cvpr/0005KZW021,DBLP:conf/nips/SaitoKSS20}. These methods require access to both source and target data, making them impractical in real-world scenarios where source data is inaccessible due to privacy or security concerns. To address this issue, source-free universal domain adaptation methods~\cite{DBLP:conf/cvpr/QuZRL0TJ23,DBLP:journals/tip/FengXT21,DBLP:journals/corr/abs-2112-08553, qu2024lead} have been developed, which only require target data for adaptation.
Unlike universal domain adaptation, which requires access to entire batches of data, OWTTA handles a more challenging scenario. In OWTTA, while all test data can be processed at once, inference and adaptation must be performed instantly with new incoming batches.

{\textbf{Test-Time Augmentation.}
Test-time augmentation improves prediction robustness by generating multiple augmented views of each test sample and aggregating their predictions~\cite{DBLP:conf/iccv/ShanmugamBBG21,DBLP:conf/nips/KimKK20,DBLP:journals/corr/abs-2507-01347}. These methods mainly exploit input-level diversity to obtain more stable predictions.
Test-time augmentation improves generalization under domain shifts by leveraging input-level diversity, whereas OWTTA focusing on adapting the model to a specific target domain in the presence of OOD interference.}

\vspace{-5pt}
\section{Methodology}~\label{sec:method}
{
In this section, we propose Reliable Neural Collapse approximation~(ReNC) for open-world test-time adaptation. The key motivation of ReNC is to exploit neural collapse as a structural prior for reliable target-domain adaptation. Specifically, we theoretically justify that, when a well-trained source model exhibits an NC-like geometry, its classifier weights can be interpreted as class prototypes. Based on this prior, ReNC first identifies reliable target samples according to their distances to the nearest prototypes, thereby filtering out OOD samples for reliable updates. Then, it progressively refines the source-domain prototypes with reliable target features and updates the model parameters under a geometry-preserving constraint in an alternating manner, encouraging the target representation to evolve toward an NC-like structure. In this way, ReNC effectively adapts the model to the target domain while maintaining a compact, balanced, and well-structured representation space.
}

\begin{figure}[t]
\centering   
\subfloat[CIFAR10-C \& Noise]{      
\includegraphics[width=0.3\columnwidth]{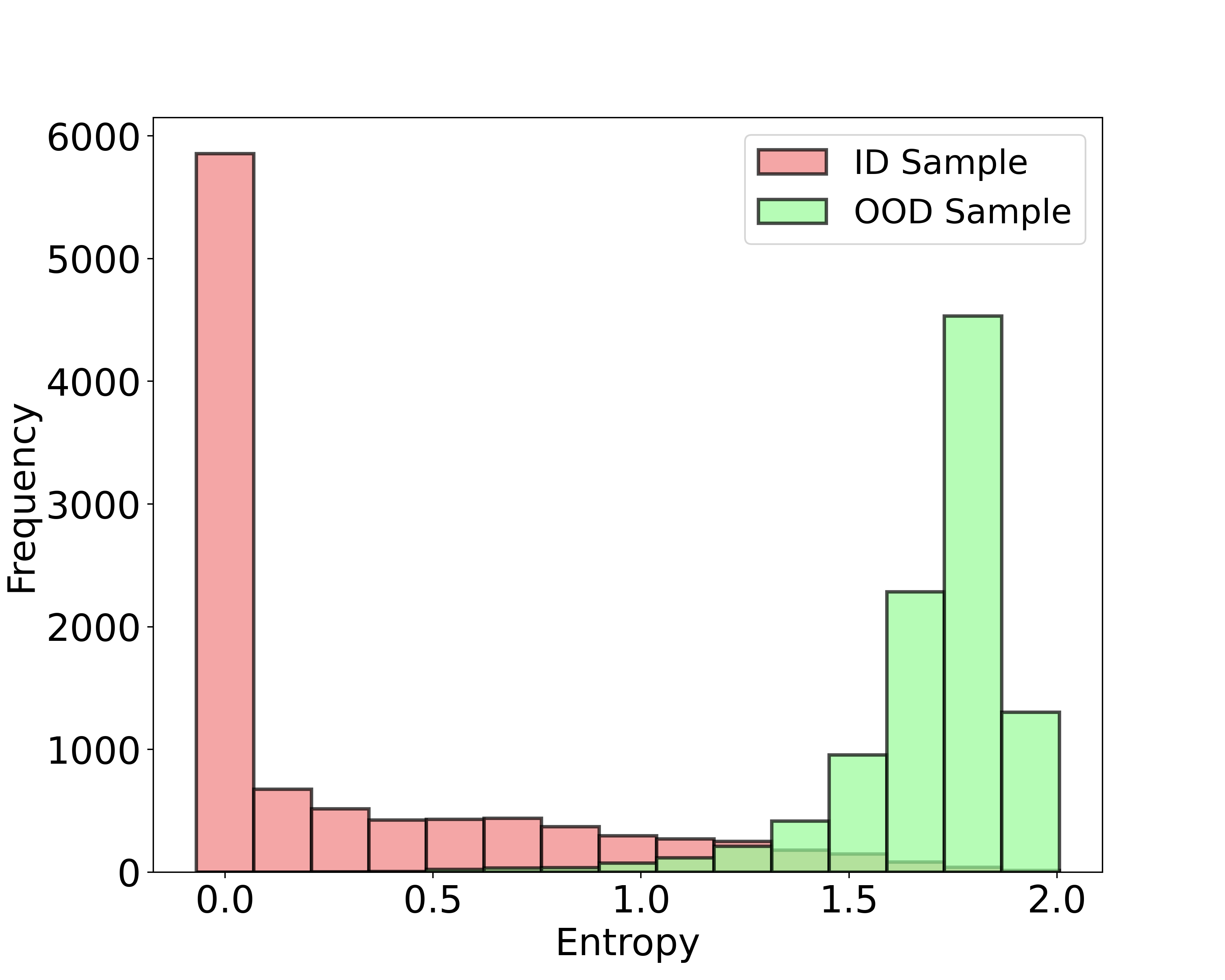}     
}
\subfloat[CIFAR10-C \& SVHN]{      
\includegraphics[width=0.3\columnwidth]{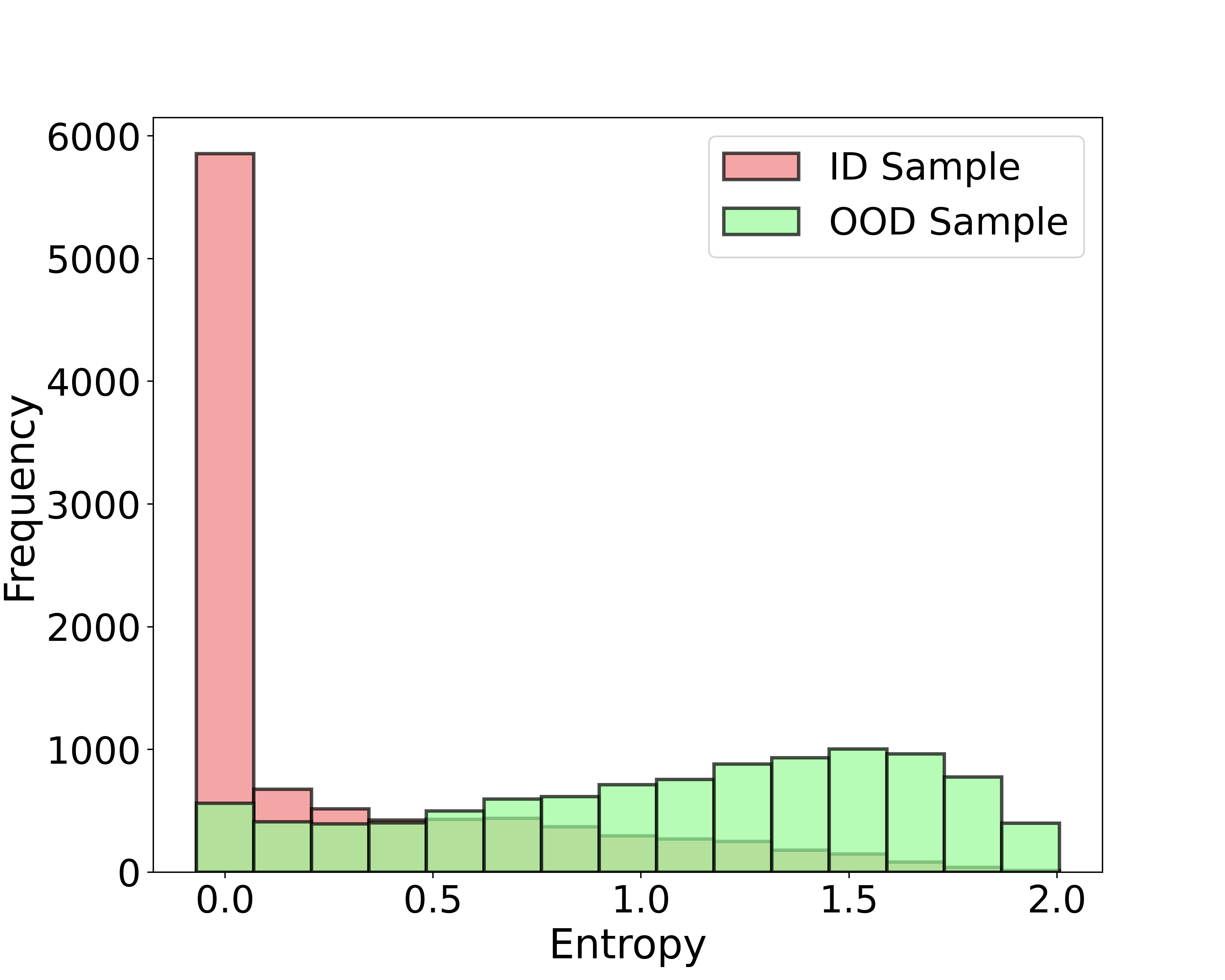}     
}
\caption{Entropy of CIFAR10-C with OOD samples ((a) Gaussian noise and (b) SVHN)  pre-trained on CIFAR10. Minimizing entropy facilitates adaptation to the target domain. However, indiscriminately minimizing all entropy in an open-world scenario would degrade the model performance.}
\vspace{-0.4cm}
\label{fig:entropy}
\end{figure}

We start by developing a simple, reliable update rule and then progressively refine our approach to handle more practical scenarios in an open-world setting.
\subsection{Reliable Update by Filtering Out OOD Samples}~\label{sec:ood}
Given a model pre-trained on the source domain $F\left(\bm{\theta}, \cdot \right)$, its logits $\mathbf{p}_i^t$ can be represented by:
\begin{equation*}
    \begin{aligned}
\mathbf{p}_i^t=F\left(\bm{\theta},\mathbf{x}_i^t\right)=\sigma\left(\mathbf{w} f\left(\bm{\theta},\mathbf{x}_i^t \right)  + \mathbf{b}\right) = \sigma\left(\mathbf{w}  \mathbf{z}_i^t + \mathbf{b}\right),
    \end{aligned}
\end{equation*}
where $\mathbf{x}_i^t$ is the input at the $t$-th batch, $f(\bm{\theta}, \cdot)$ is the encoder, $\mathbf{z}^t_i$ is the feature, $\sigma(\cdot)$ is the activation function, and $\mathbf{w}$ and $\mathbf{b}$ denote the classifier weights and bias, respectively.

Existing closed-world TTA methods assume that, despite the presence of data distribution shifts, the entropy values~\cite{DBLP:conf/icml/BelghaziBROBHC18} for target domain samples corresponding to each class remain relatively low during inferring. Based on this assumption, a straightforward way to adapt $F(\bm{\theta},\cdot)$ to the target domain $\mathcal{D}_T$ is to minimize the entropy of each logit:
\begin{equation}
    \mathcal{L}_e\left( \bm{\theta}\right)=\frac{1}{N}\sum_{i=1}^{N}
    H\left(\mathbf{p}_i^t\right),
    \label{eq:entropy}
\end{equation}
where $H(\cdot)$ is the entropy function.

However, in the open-world scenario, label distribution shift exists in test samples. As shown in Figure~\ref{fig:entropy}, the entropy of ID and OOD samples are mixed together, complicating the process of minimizing entropy.
Since the goal of minimizing entropy is to encourage each logit to form a sharp (one-hot) distribution, simply reducing the entropy of all test samples can disrupt the adaptation process.

\begin{figure} \centering    
\subfloat[CIFAR10-C\&Noise]{  
\includegraphics[width=0.3\columnwidth]{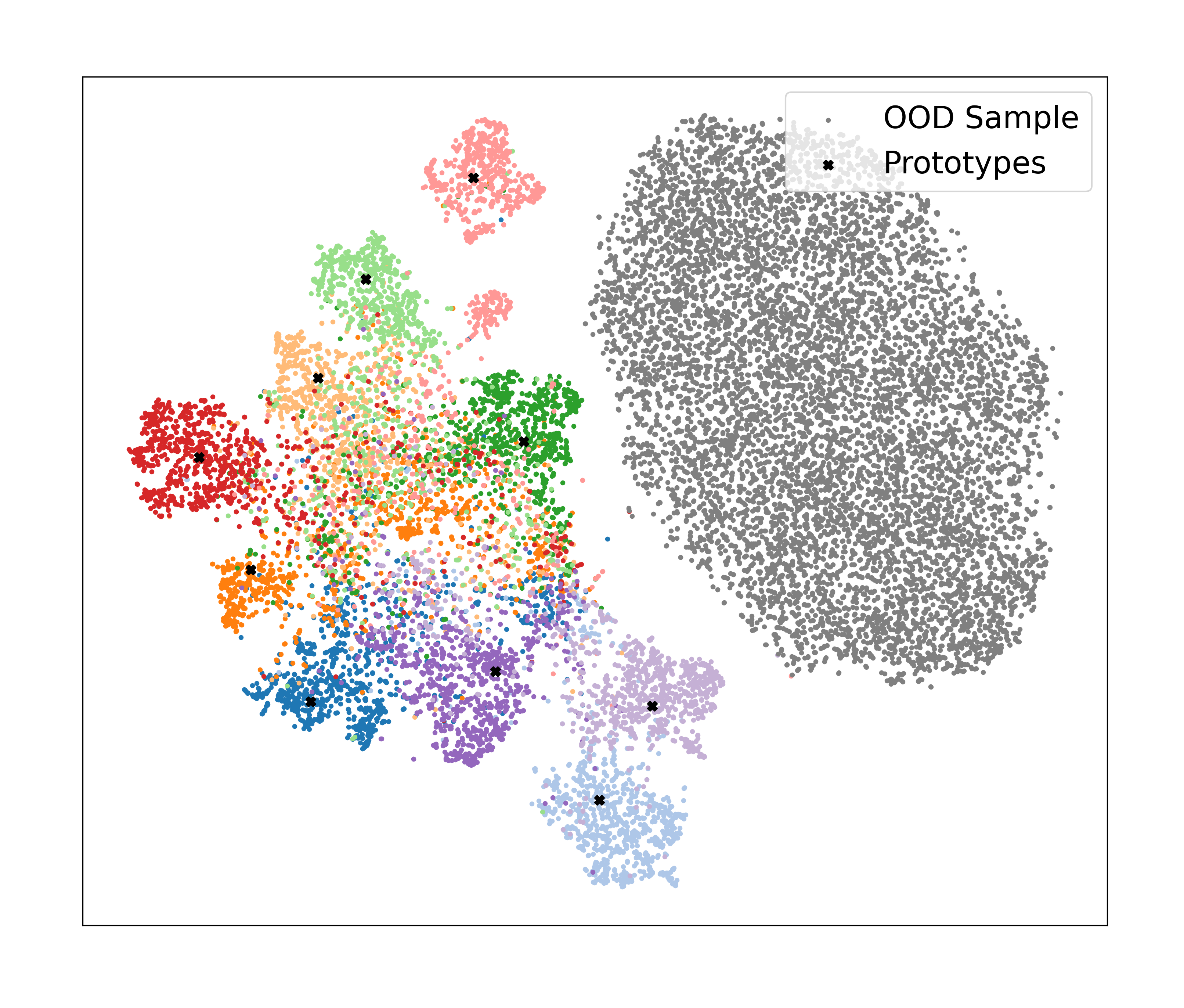}  
}  
\subfloat[CIFAR10-C\&SVHN]{      
\includegraphics[width=0.3\columnwidth]{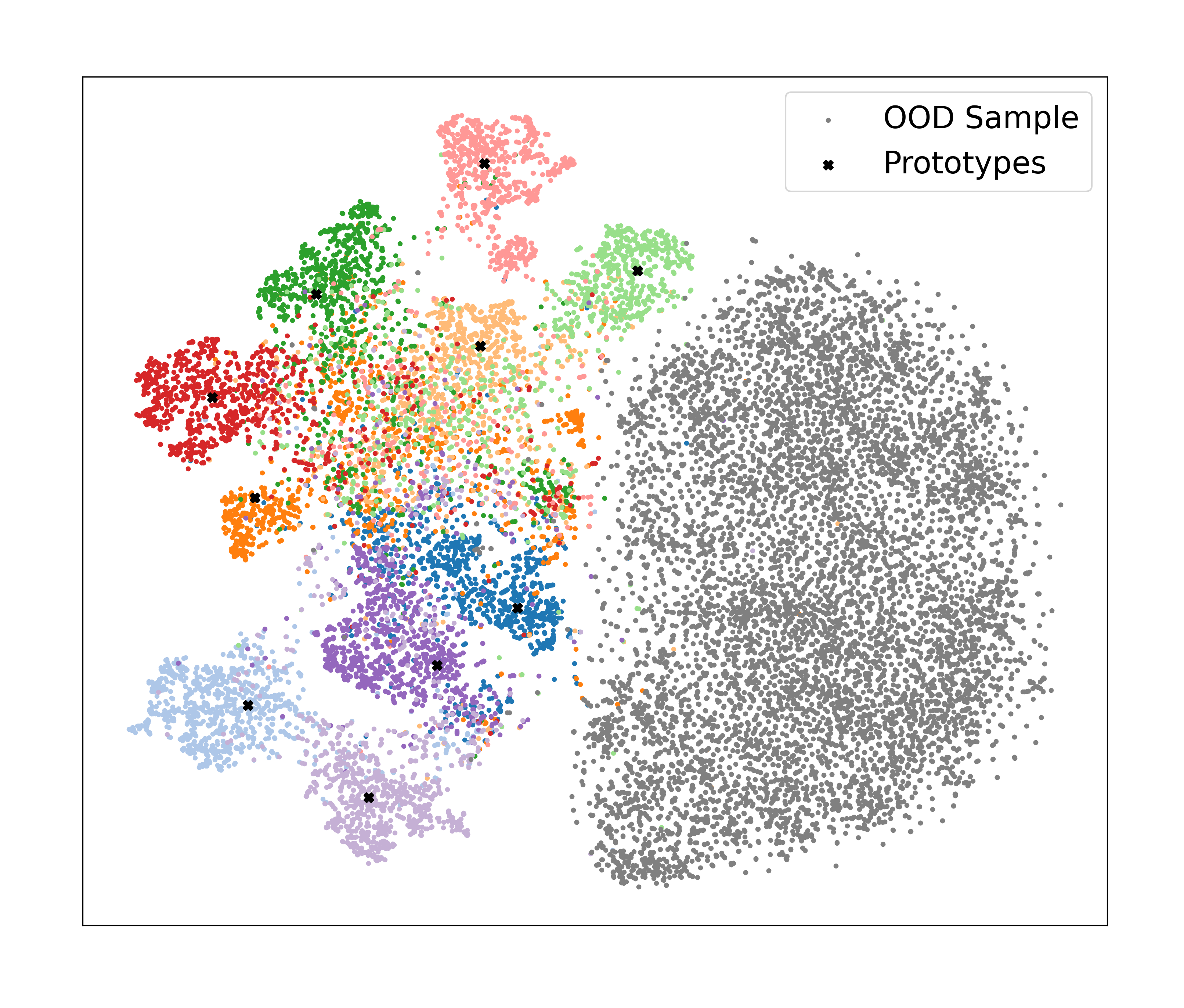}     
}
\caption{t-SNE visualization of extracted features on CIFAR10-C with OOD samples ((a) Gaussian noise and (b) SVHN) pre-trained on CIFAR10. Prototypes are generated using the feature class-means of source samples. }
\vspace{-0.4cm}
\label{fig:partition}
\end{figure}

\begin{figure}[htpb!] \centering    
\subfloat[CIFAR10-C \& Noise]{  
\includegraphics[width=0.3\columnwidth]{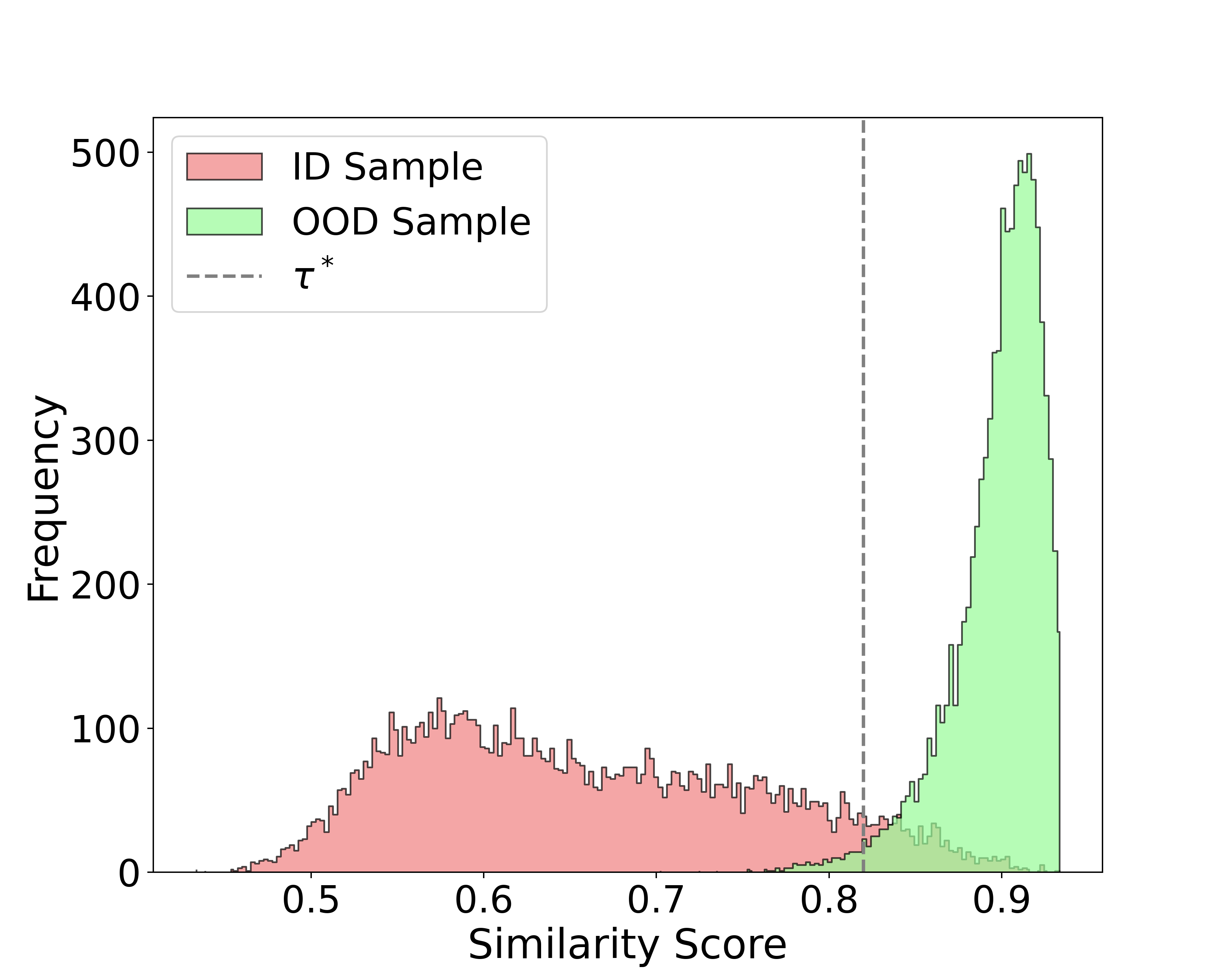}  
}  
\subfloat[CIFAR10-C \& SVHN]{  
\includegraphics[width=0.3\columnwidth]{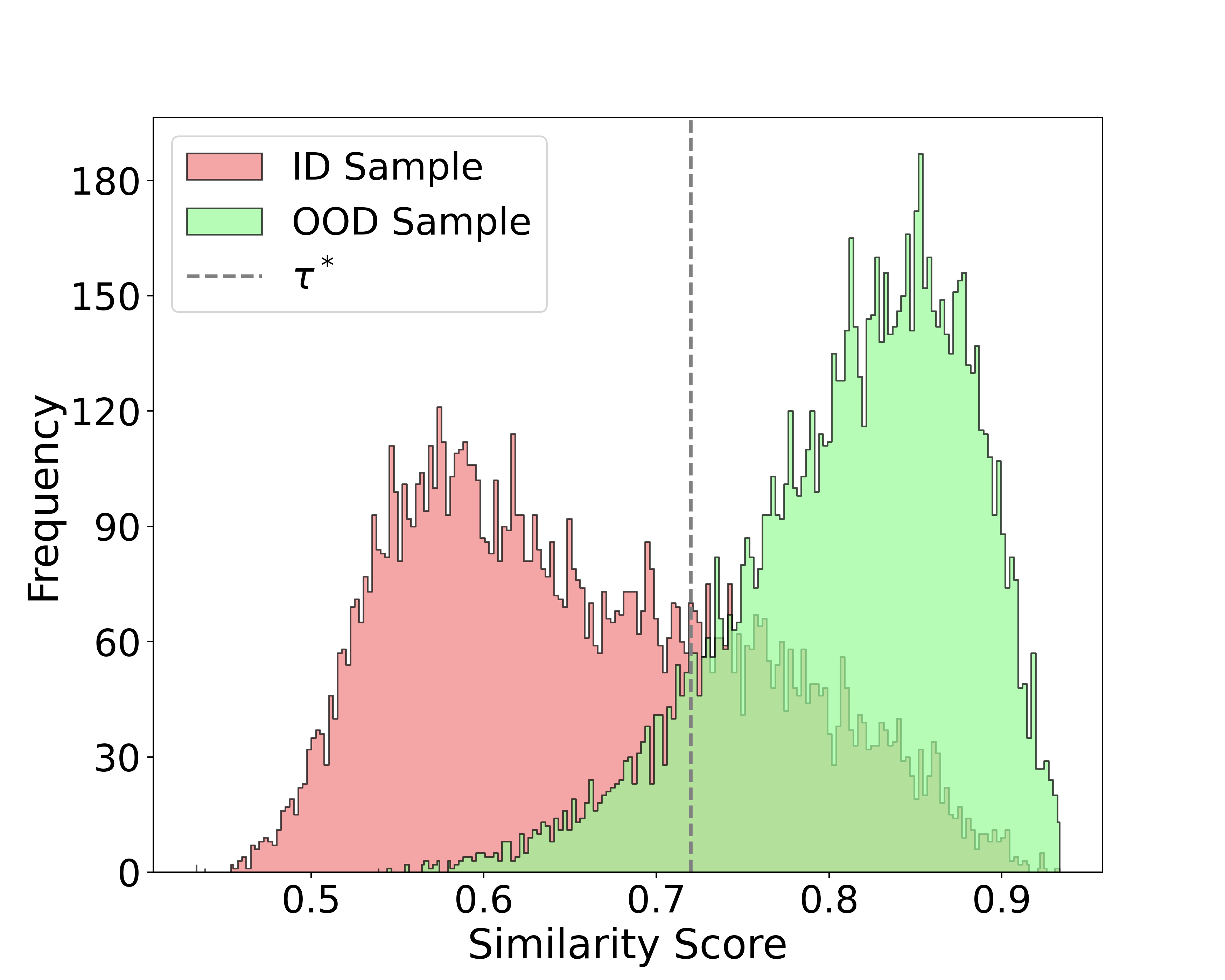}     
}
\caption{Similarity scores of CIFAR10-C with OOD samples ((a) Gaussian noise and (b) SVHN) pre-trained on CIFAR10. The differences in the similarity scores between ID and OOD samples provide a basis for their discrimination.}
\label{fig:ss}
\end{figure}


To enhance the reliability of the update process, we investigate a selective updating strategy to split the ID and OOD samples. 
{As shown in Figure~\ref{fig:partition}, ID and OOD samples exhibit distinct feature distributions, where ID samples tend to cluster around their corresponding class prototypes, computed by averaging the samples in each class.}
Inspired by few-shot learning~\cite{DBLP:conf/nips/SnellSZ17}, we aim to transform the problem of distinguishing between ID and OOD samples into a two-cluster clustering problem. To achieve this, we construct a similarity-based measure using multiple prototypes to differentiate between ID and OOD samples.

 Specifically, given $K$ prototypes from source domain as $\mathbf{C}=\left\{\mathbf{c}_j\right\}_{j=1}^{K}$, where $\mathbf{c}_j$ is calculated by the mean value of the features from $j$-th class, the similarity score between the test sample and its nearest prototype is:
\begin{equation}
    s^t_i = 1-\underset{\mathbf{c}_j\in \mathbf{C}}{\max}\frac{\mathbf{z}_i^{t^\top} \mathbf{c}_j}{\|\mathbf{z}_i^t\| \|\mathbf{c}_j\|}.
    \label{eq:ss_socre}
\end{equation}

Some existing method~\cite{gao2024unified} utilize a symmetric Gaussian Mixture Models (GMMs) with two components to fit ID and OOD samples based on their similarity scores. However, it is inappropriate because each component does not adhere to a symmetric Gaussian distribution, as shown in Figure~\ref{fig:ss}. 
Instead, we approach this problem by learning an adaptive threshold~$\tau_t$ to partition ID and OOD samples, effectively treating it as a clustering task:
\begin{equation}
    \begin{aligned}
        \underset{\tau_t}{\min}~&\frac{1}{N^+} \sum_i \Big[ s_i^t - \frac{1}{N^+} \sum_j \mathbf{1}(s_j^t \leq \tau_t) s_j^t \Big]^2 \\
        &+ \frac{1}{N^-} \sum_i \Big[ s_i^t - \frac{1}{N^-} \sum_j \mathbf{1}(s_j^t > \tau_t) s_j^t \Big]^2,
    \end{aligned}
    \label{eq:ood}
\end{equation}
where $N^+ = \sum_i \mathbf{1}\left(s^t_i \leq \tau_t\right)$ and $N^- = \sum_i \mathbf{1}\left(s_i^t > \tau_t\right)$ represent the number of ID and OOD samples respectively.
Optimizing Eq.~\eqref{eq:ood} enables finding an optimal threshold $\tau_t^*$ to minimize the intra-cluster variations.
Then, the objective function can be revised into:
\begin{equation}
    \mathcal{L}_e^+\left( \bm{\theta},\mathbf{C}\right)=
    \frac{1}{N^+}\sum_{i=1}^{N^+}
    H\left(\mathbf{p}_i^t\right).
    \label{eq:entropy+}
\end{equation}

By applying Eq.~\eqref{eq:entropy+}, reliable updates can be achieved by minimizing the entropy of $N^+$ samples. 

{In this process, the prototypes $\mathbf{C}$ serve as reliable class anchors for distinguishing ID samples from OOD ones, and thus are crucial for stable target-domain updates.}

\begin{figure*} \centering    
\subfloat[$K=2$]{  
\includegraphics[width=0.3\columnwidth]{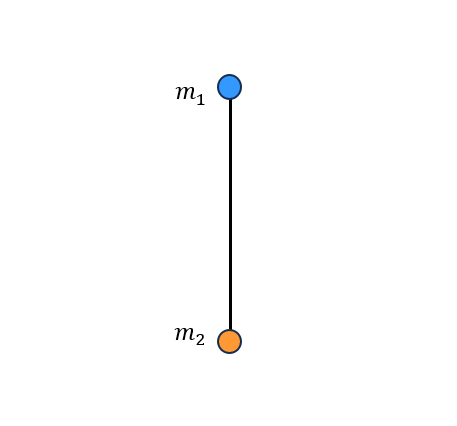}  
}  
\subfloat[$K=3$]{  
\includegraphics[width=0.3\columnwidth]{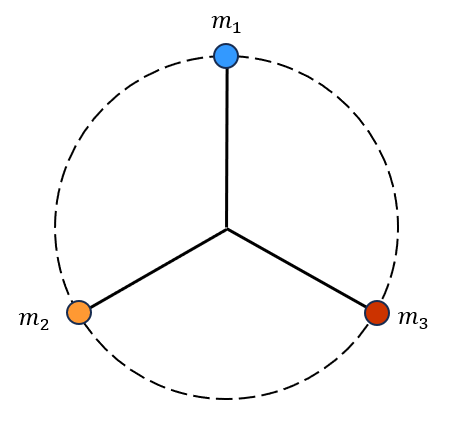}  
}     
\subfloat[$K=4$]{      
\includegraphics[width=0.3\columnwidth]{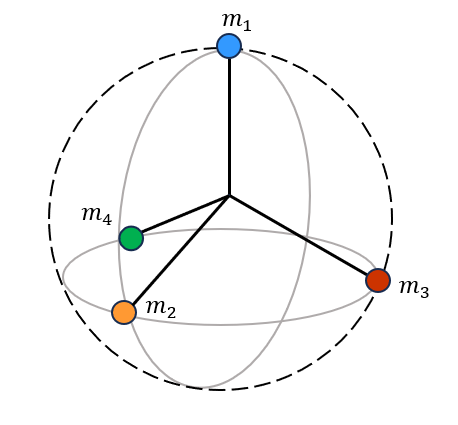}     
}
\caption{Simplex Equiangular Tight Frame~(Simplex ETF) with different class numbers.}
\label{fig:Simplex_etf}
\end{figure*}

\subsection{Source-Free Update with Pre-trained Classifier Weights}
\label{Sec:souce-free}




{As discussed in Subsection~\ref{sec:ood}, reliable updates rely on access to class prototypes. However, in Open-World Test-Time Adaptation~(OWTTA), such prototype information is unavailable, since only a pre-trained model is given and unlabeled target samples arrive batch-by-batch. Therefore, we derive the prototypes directly from the pre-trained model itself.}

{Recent neural collapse theory~\cite{DBLP:conf/nips/ZhuDZLYSQ21} provides a principled explanation for the final state of supervised training. Specifically, when a sufficiently large neural network is well-trained on the training data, it tends to exhibit the following properties.
}


\begin{myTheo}[Convergence to Simplex ETF~\cite{DBLP:conf/nips/ZhuDZLYSQ21}] For a sufficiently large neural network, the class-means of the embeddings centered at the global mean become both linearly separable and maximally distant when the model converges. These class-means lie on a sphere centered at the origin, forming a Simplex ETF:
\begin{equation*}
\frac{\langle \mathbf{c}_i - \mathbf{c}_G, \mathbf{c}_{j} - \mathbf{c}_G \rangle}{\|\mathbf{c}_i - \mathbf{c}_G\|_2 \|\mathbf{c}_{j} - \mathbf{c}_G\|_2} {=}
\begin{cases} 
1, & i = j \\
-\frac{1}{K-1}, & i \neq j, 
\end{cases},
\end{equation*}
\begin{equation*}
\|\mathbf{c}_i - \mathbf{c}_G\|_2 - \|\mathbf{c}_{j} - \mathbf{c}_G\|_2 {=}0, \quad \forall 1\leq i,j \leq K,
\end{equation*}
\label{Theo:NC2}
where $\mathbf{c}_j = \frac{\sum_{i=1}^{N} \mathbf{z}_i \mathbb{I}({\Psi}(\mathbf{z}_i) = j)}{\sum_{i=1}^{N} \mathbb{I}({\Psi}(\mathbf{z}_i) = j)}
$ is the class-mean of $j$-th class and $\mathbf{c}_G=\frac{1}{N}\sum_{i=1}^N \mathbf{z}_{i}$ is the global mean of feature, {both are computed under the neural collapse state}. \hfill\rule{2mm}{2mm}
\end{myTheo}

\begin{myTheo}[Convergence to Self-duality~\cite{DBLP:conf/nips/ZhuDZLYSQ21}] For a sufficiently large neural network, the last-layer linear classifiers reside in the dual vector space of the class-means when the model has converged:
    \begin{equation}    
        \mathcal{N}C_3 := \left\| \frac{\mathbf{W} \overline{\mathbf{C}}}{\left\| \mathbf{W} \overline{\mathbf{C}} \right\|_F} - \frac{1}{\sqrt{K-1}} \left( \mathbf{I}_K - \frac{1}{K} \mathbf{\mathbf{1}}_K \mathbf{\mathbf{1}}_K^\top \right) \right\|_F,
    \end{equation}
    \label{theo:NC3}
where $\overline{\mathbf{C}} = [\mathbf{c}_1-\mathbf{c}_G, \dots,\mathbf{c}_K-\mathbf{c}_G]$ represents the centered class-mean matrix.
\hfill\rule{2mm}{2mm} 
\end{myTheo}

\begin{myTheo}[Variability Collapse~\cite{DBLP:conf/nips/ZhuDZLYSQ21}]
As training progresses, the features within each class converge to their respective class-means, causing within-class variation to collapse to zero. Let the within-class covariance matrix be denoted as $\Sigma_\mathbf{W} \in \mathbb{R}^{d \times d}$ and the between-class covariance matrix as~$\Sigma_\mathbf{B} \in \mathbb{R}^{d \times d}$, defined as follows:
\begin{equation*}
    \begin{aligned}
        &\Sigma_\mathbf{W} := \frac{1}{NK}\sum_{i=1}^{N}\sum_{j=1}^{K}\mathbf{d}_{ij}\mathbf{d}_{ij}^{T},\\ 
        &\Sigma_\mathbf{B} := \frac{1}{K}\sum_{j=1}^{K}\left( \mathbf{c}_j-\mathbf{c}_G\right)\left( \mathbf{c}_j-\mathbf{c}_G\right)^{T},
    \end{aligned}
\end{equation*}
where $\mathbf{d}_{ij}= \mathbb{I}({\Psi}(\mathbf{z}_i) = j)\cdot(\mathbf{z}_i-\mathbf{c}_j)$ is the distance between the $i$-th feature and its class-mean. Then, we can measure the extent of within-class variability collapse by analyzing the relative magnitudes of $\Sigma_\mathbf{B}$ and $\Sigma_\mathbf{W}$ for the learned features:
    \begin{equation}
\mathcal{NC}_1:=\Sigma^{\dagger}_\mathbf{B} \Sigma_\mathbf{W} \to 0,
\end{equation}
\label{theo:NC1}
where $\dagger$ denotes the pseudo-inverse.
\hfill\rule{2mm}{2mm} \end{myTheo}

{These theorems indicate that, at convergence, both the class means and classifier weights tend to align with a Simplex ETF structure, as depicted in Figure~\ref{fig:Simplex_etf}. Since OWTTA starts from a pre-trained source model, we assume that the source model has been sufficiently trained and thus approximately satisfies this neural collapse property. See Table~\ref{tab:nc_original} for empirical verification.}

\begin{assumption}
    The pre-trained model can be considered well-trained on the source domain, with its {classifier} weights collapsing into a Simplex ETF.
    \label{assum:well-train}
\hfill\rule{2mm}{2mm} 
\end{assumption}

\begin{assumption}~\label{assum:neglect}
Let $F(\bm{\theta}^*,\cdot)=\sigma\bigl(\mathbf{w}^*f(\bm{\theta}^*,\cdot)+\mathbf{b}^*\bigr)$
be a classification network that has reached the neural collapse state, where $\mathbf{w}^*$ and $\mathbf{b}^*$ are its classifier weights and bias, respectively. Let $\mathbf{z}=f(\bm{\theta}^*,\mathbf{x})$ denote the feature representations of the training set $\mathcal{S}_S\in \{(\mathbf{x}_i,y_i)|\mathbf{x}_i \in \mathcal{D}_S\}$. Then, for some sufficiently small positive real $\epsilon$, it holds that 
    \begin{equation}
       \frac{\|\mathbf{b}_j^* \|}{\langle \mathbf{w}^*_{:,j},\mathbf{z}_{\Psi({\mathbf{z})=j}}\rangle}<\epsilon,\quad\forall j.
   \end{equation}
\end{assumption}

{Under Assumption~\ref{assum:neglect}, the optimal bias $\mathbf b^*$ is negligible in the neural collapse state. See Table~\ref{sec:NC_} for empirical verification. Accordingly, the bias term can be omitted, and the optimization problem
$\underset{\mathbf{w}_{:,j},\mathbf{b}_j}{\max}\,
\langle\mathbf{w}_{:,j},\mathbf{z}_{\Psi(\mathbf{z})=j}\rangle+\mathbf{b}_j$
reduces to
$\underset{\mathbf{w}_{:,j}}{\max}\,
\langle\mathbf{w}_{:,j},\mathbf{z}_{\Psi(\mathbf{z})=j}\rangle$.
This indicates that the classifier weight of each class is optimized toward the same direction as its corresponding class prototype. We formalize this relation in the following lemma.}


\begin{lemma}\label{lemma:self-dual}
    {Let $F(\bm{\theta}^*,\cdot)=\sigma\bigl(\mathbf{w}^*f(\bm{\theta}^*,\cdot)+\mathbf{b}^*\bigr)$ be a sufficiently trained classification network that approximately satisfies the neural collapse property. Let $f(\bm{\theta}^*,\mathbf{x})$ denote the feature representation of a source training sample $\mathbf{x}$ from $\mathcal{S}_S=\{(\mathbf{x}_i,y_i)\mid \mathbf{x}_i \in \mathcal{D}_S\}$.} Let $\mathbf{c}_j$ be the prototype of the $j$-th class. Then we have:
    \begin{equation}
        \frac{\mathbf{c}_j}{\|\mathbf{c}_j\|}=\frac{\mathbf{w}^*_{:,j}}{\|\mathbf{w}^*_{:,j}\|}, \quad\forall j.
    \end{equation}
\end{lemma}

{Following Lemma~\ref{lemma:self-dual}, we use the normalized classifier weights as class prototypes under the source-free setting. Therefore, Eq.~\eqref{eq:ood} can be computed at each batch step without accessing source data, enabling reliable updates with selected target samples.}

\subsection{Neural Collapse Approximation}
{We further consider the effect of data distribution shifts. The neural collapse structure learned by the source model reflects the source-domain feature geometry, but may become misaligned with target-domain representations due to the data distribution shifts. From the neural collapse perspective, this feature-prototype misalignment between shifted target representations and source-domain prototypes becomes a key reason for performance degradation.}


{
\begin{myProp}\label{proposition:1}
Given a source domain training set~$\mathcal{S}_S\in \{(\mathbf{x}_i,y_i)|\mathbf{x}_i \in \mathcal{D}_S\}$ and a target domain test set~$\mathcal{S}_T\in \{\mathbf{x}_i|\mathbf{x}_i \in \mathcal{D}_T\}$. Under the neural collapse state, let~$\mathbf{c}^*_S$ and $\mathbf{c}^*_T$ denote the prototypes of source and target domains, respectively, each forming a Simplex ETF:
\begin{equation*}
\begin{aligned}
    &\mathbf{c}_S^*\,\mathbf{c}_S^{*\top}
=\alpha\Bigl(\mathbf I_K - \tfrac1K\,\mathbf 1_K\mathbf 1_K^\top\Bigr),
\mathbf{c}_T^*\,\mathbf{c}_T^{*\top}
=\beta\Bigl(\mathbf I_K - \tfrac1K\,\mathbf 1_K\mathbf 1_K^\top\Bigr),
\end{aligned}
\end{equation*}
where $\alpha$ and $\beta$ denote the respective scaling factors for the source and target Simplex ETFs.
\end{myProp}
}
{
By Proposition~\ref{proposition:1}, if $\mathcal{D}_S\neq \mathcal{D}_T$, then $\mathbf{c}^*_S\neq\mathbf{c}^*_T$, i.e., the source and target prototypes differ due to the data distribution shift, which leads to classification errors.
{Based on this understanding, we aim to guide target representations toward a target-domain NC-inspired geometry during adaptation.}

{
\begin{lemma}~\label{lemma:label}
Let~$F(\bm{\theta}^*,\cdot)=\sigma\bigl(\mathbf{w}^*f(\bm{\theta}^*,\cdot)+\mathbf{b}^*\bigr)$ be a classification network that has reached the neural collapse state with the training set~$\mathcal{S}_S\in \{(\mathbf{x}_i,y_i)|\mathbf{x}_i \in \mathcal{D}_S\}$. Then, for each training sample $\mathbf{x}_i$, its logits vector~$\mathbf{p}=F(\bm{\theta}^*,\mathbf{x}_i)$ is a one-hot vector, i.e.,
\begin{equation*}
\mathbf{{p}}_j = 
\begin{cases} 
1, & \Psi(\mathbf{x}_i) = j, \\
0, &  \text{otherwise}.
\end{cases}
\end{equation*}
\end{lemma}}

{As stated in Lemma~\ref{lemma:label}, the logits of a classification network tend to degenerate into one-hot vectors under the neural collapse state. The entropy loss in Eq.~\eqref{eq:entropy+} encourages confident predictions, which implicitly pushes each sample toward its corresponding class mean and promotes a more compact target-domain representation. This connection to neural collapse explains why entropy minimization can improve adaptation effectiveness in TTA.}


{However, as indicated by Proposition~\ref{proposition:1}, distribution shifts may alter the target-domain feature geometry, making direct entropy minimization in Eq.~\eqref{eq:entropy+} blindly push target features toward source-domain prototypes. Such indiscriminate updates may distort the embedding space and damage the structured representations required for reliable adaptation.}

{We therefore seek to approximate the neural collapse state in the target domain. Since exact neural collapse is difficult to achieve in the unsupervised setting, we use it as a reliable structural prior for target adaptation. Specifically, we jointly update the model parameters and prototypes, and further leverage embedding-structure maintenance and stochastic prototype updates to promote balanced and compact representations during adaptation.}

\subsubsection{Maintain Embedding Structure}\label{sec:embedding}
{According to neural collapse theory, learned class representations tend to form a balanced geometric structure, i.e., a Simplex ETF, at convergence. We regard this property as a structural prior and preserve the balanced latent embedding geometry during target adaptation. Let $\mathbf{w}^*$ denote the classifier weights, and let $\mathbf{z}_{\Psi(\mathbf{z})=j}$ denote the feature representations assigned to the $j$-th class.} Then we have:
\begin{equation*}
\begin{aligned}
    \hat{\mathbf{p}}^* &= \frac{1}{K}\sum_{j=1}^K \sigma(\langle\mathbf{w}^*,\mathbf{z}_{\Psi(\mathbf{z})=j}\rangle)=\frac{1}{K}\sum_{j=1}^K\sigma(\langle\mathbf{w}^*,\mathbf{c}_j\rangle)\\
    &=\frac{1}{K}\sum_{j=1}^K\sigma([-\frac{1}{K-1},-\frac{1}{K-1},\dots,1,\dots,-\frac{1}{K-1}])\\
    &=\frac{1}{K}\mathbf{1}_K,
\end{aligned}
\end{equation*}
{where $\hat{\mathbf{p}}^*$ denotes the ideal class-averaged prediction under neural collapse, corresponding to a uniform distribution over classes.}

{Guided by this uniform prior, we encourage the average logits of ID samples to approach $\frac{1}{K}\mathbf{1}_K$, thereby maintaining a balanced embedding structure during adaptation:}
\begin{equation}\label{eq:balance}
   \mathcal{L}_{d}\left(\bm{\theta}\right)= D_{KL}\left( {\hat{\mathbf{p}}}^t \Big\| \frac{1}{K}\mathbf{1}_K\right),
\end{equation}
where $\hat{\mathbf{p}}^t=\frac{1}{N_+}\sum_{i=1}^{N^+}\mathbf{p}_i^t$ is the average of logits of $t$-th batch, $D_{KL}$ is the KL-divergence, ${\mathbf{p}}^t$ is the output of ID samples.
{Eq.~\eqref{eq:balance} regularizes the adaptation by encouraging balanced class-wise predictions, which facilitates the preservation of the neural collapse-inspired embedding structure.}

\subsubsection{{Stochastic Prototype Update}}\label{sec:prototype_update}
{As stated in Proposition~\ref{proposition:1}, source-domain and target-domain prototypes may differ under distribution shifts. Therefore, we further investigate how to update prototypes during adaptation. In OWTTA, target samples arrive in a single-pass stream, and each mini-batch usually covers only a subset of classes. This naturally introduces a class-coverage imbalance during prototype updates. For example, when the number of classes is much larger than the batch size, many classes are absent from the current batch. Updating prototypes only with the current batch may therefore produce biased estimates of the target distribution, hindering reliable adaptation and degrading OWTTA performance.}


{To address this issue, we propose a stochastic prototype update mechanism that selectively updates only the prototypes corresponding to classes present in the current batch, while keeping others fixed. This strategy ensures that updates are more relevant to the incoming data and improves adaptation efficiency.} To facilitate this, the objective function is extended as follows:
\begin{equation}
\begin{aligned}
    \mathcal{L}\left( \bm{\theta},\mathbf{C}\right)=&~\mathcal{L}_e^+\left(\bm{\theta}, \mathbf{C} \right)  + \lambda\mathcal{L}_d\left( \bm{\theta}\right) + \mathcal{L}_s\left( \bm{\theta}, \mathbf{C}\right)\\
    =&~\frac{1}{N^+}\sum_{i=1}^{N^+}
    H\left(\mathbf{p}_i^t\right){+\lambda D_{KL}\left( {\mathbf{p}}^t \Big\| \frac{1}{K}\mathbf{1}_K\right)} \\
    &~+\frac{1}{N^+}\sum_{i=1}^{N^+}\sum_{j=1}^{K}\xi_{ij}^{t} \left\| \mathbf{z}_i^{t}-\mathbf{c}_j^{t}\right\|_{2}^{2},
\end{aligned}
\label{eq:obj_wo_average}
\end{equation}
where $\lambda$ is the trade-off parameter and $\xi\in \left\{ 0,1\right\}$ is an indicator factor assigning $\mathbf{z}^t_{i}$ into its nearest prototype:
\begin{equation}
    \begin{aligned}
\xi_{ij}^t &= 
\begin{cases} 
1, & \text{if } j = {\underset{j}{\arg\max}~\frac{\exp \left(
        {-\left\|\mathbf{z}_i^t - \mathbf{c}^{t-1}_j\right\|_2^2}
        \right)}
        {\sum_j^{K} \exp\left(-\left\|\mathbf{z}_i^t- \mathbf{c}^{t-1}_j\right\|_2^2\right)}},\\
0, & \text{otherwise}.
\end{cases}
\\
\end{aligned} 
\end{equation}


{The first term of Eq.~\eqref{eq:obj_wo_average} is guided by the compactness prior in Theorem~\ref{theo:NC1}, encouraging each sample to move closer to its corresponding prototype. The second term is inspired by the simplex-ETF geometry in Theorem~\ref{Theo:NC2}, encouraging a well-structured inter-class embedding space. The third term alleviates the prototype mismatch described in Proposition~\ref{proposition:1} by progressively adapting the prototypes toward the target-domain feature distribution. These three terms jointly incorporate the neural collapse prior into unsupervised target adaptation, enabling the model to form a compact and balanced target-domain representation without requiring label supervision.}

 \subsubsection{Alternating Optimization}
  {Unlike the network parameters~$\bm{\theta}$, prototypes $\mathbf{C}$ can be unstable when directly updated by stochastic gradients.}
 This instability occurs because the prototypes estimated from different mini-batches can vary significantly. To achieve the stochastic prototype update, we propose a decomposition-coordination optimization approach based on the Alternating Direction Method of Multipliers (ADMM) to optimize Eq.~\eqref{eq:obj_wo_average}. In particular, we employ an auxiliary variable $\bm{\mu}$ to alternately update the model parameters and prototypes:
\begin{equation*}
    \begin{aligned}
        \mathcal{L}\left( \bm{\theta}, \bm{\mu} , \mathbf{C}\right)=&~
        \mathcal{L}_e^+\left(\bm{\theta}, \bm{\mu} \right) + \lambda\mathcal{L}_d\left( \bm{\theta}\right)+\mathcal{L}_s\left( \bm{\theta}, \bm{\mu}\right)\\
        &~+\rho \sum_{j=1}^{K} \left\|\bm{\mu}_j-\mathbf{c}_j\right\|_2^2,
    \end{aligned}
\end{equation*}
where $\rho$ is a trade-off parameter controlling the prototype updates between batches.
Our decomposition-coordination optimization technique consists of three iterative steps:
\begin{equation*}
    \begin{aligned}
        &\bm{\theta}^t=\underset{\bm{\theta}}{\arg \min} ~\mathcal{L}\left( \bm{\theta},\bm{\mu}^{t-1},\mathbf{C}^{t-1} \right),\\
        &\bm{\mu}^t=\underset{\bm{\mu}}{\arg \min} ~\mathcal{L}\left( \bm{\theta}^t,\bm{\mu},\mathbf{C}^{t-1} \right),\\
        &\mathbf{C}^t=\underset{\mathbf{C}}{\arg \min} ~\mathcal{L}\left( \bm{\theta}^t,\bm{\mu}^{t},\mathbf{C} \right).
    \end{aligned}
\end{equation*}

We then develop efficient update solutions for each sub-problem, allowing for neural collapse approximation.

\textbf{For updating $\bm{\theta}$:} Since no closed-form solution exists, we optimize $\bm{\theta}^t$ using gradient descent:
\begin{equation}
    \bm{\theta}^t=\bm{\theta}^{t-1}-\epsilon \left.\frac{\partial\mathcal{L}\left( \bm{\theta},\bm{\mu}^{t-1},\mathbf{C}^{t-1}\right)}{\partial \bm{\theta}}\right|_{\bm{\theta}=\bm{\theta}^{t-1}},
    \label{eq:theta}
\end{equation}
where $\epsilon$ is the step size. Eq.~\eqref{eq:theta} can be executed using advanced gradient descent methods.

\textbf{For updating $\bm{\mu}$:} The sub-problem involves a two-objective optimization. By setting the gradient of the objective function to zero, we can obtain an analytic solution:
\begin{equation}
\frac{\partial \mathcal{L}\left( \bm{\theta}^t,\bm{\mu},\mathbf{C}^{t-1} \right)}{\partial \bm{\mu}_j} = 0 \implies \bm{\mu}_{j}^{t} = \frac{\rho \mathbf{c}_j^{t-1} + \frac{\eta}{N^+} \sum_{i=1}^{N^+} \xi_{ij}^{t} \mathbf{z}_i^{t}}{\rho + \frac{\eta}{N^+} \sum_{i=1}^{N^+} \xi_{ij}^{t}},
 \label{eq:ema}
\end{equation}
where $\eta$ is a balance factor. The numerator of Eq.~\eqref{eq:ema} controls the sample weights of each prototype, while the denominator represents the sample counts assigned to the class. 
Let $\kappa = \frac{\rho}{\rho + \frac{\eta}{N^+} \sum_{i=1}^{N^+} \xi_{ij}^{t}} $ and 
$\bar{\mathbf{c}}_j^{t} = \frac{\sum_{i=1}^{N^+} \xi_{ij}^{t}\mathbf{z}_{i}^{t}}{\sum_{i=1}^{N^+} \xi_{ij}^{t}}$. Eq.~\eqref{eq:ema} can further simplified to the following form:
\begin{equation}
    \bm{\mu}_j^{t} = \kappa \mathbf{c}_j^{t-1} + (1 - \kappa) \bar{\mathbf{c}}_j^{t},\quad j = 1, 2, \ldots, K,
    \label{eq:ema2}
\end{equation}
where $\kappa$ acts as a self-learning parameter that balances the influence between the previous prototype $\mathbf{c}_j^{t-1}$ and the prototype calculated in the current batch, $\bar{\mathbf{c}}_j^{t}$. 
It is noteworthy that in Eq.~\eqref{eq:ema2}, $\bar{\mathbf{c}}_j^{t}$ is calculated based not only on the features assigned to the class but also on the number of samples allocated to it.
In fact, Eq.~\eqref{eq:ema2} corresponds to the exponential moving average (EMA), which has been demonstrated that the use of it ensures both stability and efficiency in scenarios of mini-batch updates~\cite{DBLP:conf/icml/XieZCC18}.

\textbf{For updating $\mathbf{C}$:} By setting the gradient of the objective function to zero, the analytic solution of $\mathbf{C}$ is:
\begin{equation}
\frac{\partial \mathcal{L} (\bm{\theta}^t,\bm{\mu}^t,\mathbf{C})}{\partial \mathbf{c}_j} = 0 \implies \mathbf{c}_j^{t} = \bm{\mu}_j^{t}, \quad j = 1, 2, \ldots, K.
\label{eq:c}
\end{equation}



{To summarize, our ReNC method filters out the OOD samples at first and approximates the neural collapse state in the target domain by updating prototypes stochastically to achieve reliable and efficient OWTTA.}
Finally, the predictions $\hat{y}_i^t$ for each batch can be obtained by:
 \begin{equation}
     \begin{aligned}
         \hat{y}_i^t =\begin{cases}F(\bm{\theta}^t,\mathbf{x}_i^t), & s_i^t < \tau_t^*,\\|\mathcal{Y}_S| + 1, & {\text{otherwise}}. \end{cases}
     \end{aligned}
     \label{eq:prediction}
 \end{equation}


A detailed overview of the proposed method is provided in Algorithm~\ref{alg:TTA}.

\begin{algorithm}[htpb!]
\caption{Reliable Neural Collapse approximation~(ReNC)}
\label{alg:TTA}
\begin{algorithmic}[1]
\REQUIRE Test samples $\mathbf{x}$ from the target domain $\mathcal{S}_T$, pre-trained model $F(\bm{\theta}, \cdot)$;
\Statex \textcolor{gray}{\# \textit{Initialize prototypes}}
\STATE \textbf{Initialization:} Extract prototypes $\mathbf{C}$ from classifier weights $\mathbf{w}$ and initialize $\bm{\mu}=\mathbf{C}$;
\FOR{$t \leftarrow 1$ to $T$}
       \Statex \textcolor{gray}{\# \textit{Partition data into ID~($N^+$) and OOD~($N^-$) samples}}
        \STATE Calculate $s_i^t$ and $\tau_t^*$ by Eq.~\eqref{eq:ss_socre} and Eq.~\eqref{eq:ood};
        \Statex \textcolor{gray}{\# \textit{Neural collapse approximation}}
        \STATE Update~$\bm{\theta}^t$ by Eq.~\eqref{eq:theta}, update~$\bm{\mu}^t$ by Eq.~\eqref{eq:ema}, update~$\mathbf{c}^t$ by Eq.~\eqref{eq:c};
         \Statex \textcolor{gray}{\# \textit{Inference}}  
        \STATE Obtain $\hat{y}_i^t$ by Eq.~\eqref{eq:prediction}.
\ENDFOR
\end{algorithmic}
\end{algorithm}

\section{Experiments}~\label{sec:experiments}
In this section, we empirically validate the proposed ReNC on several open-world benchmarks. We begin by describing the datasets, evaluation metrics, baselines.
Next, we conduct extensive experiments on five open-world benchmarks to demonstrate the effectiveness of our ReNC. {Additionally, we validate Assumption~\ref{assum:well-train} and Assumption~\ref{assum:neglect}, and further discuss the connection between TTA and neural collapse. Finally, we conduct comprehensive analyses, including ablation studies, computational cost analysis, parameter sensitivity analysis, different ID/OOD ratio settings, continual OWTTA evaluation, ViT-backbone evaluation, and significance tests, to further assess the robustness and effectiveness of our method.}

\begin{table}[htpb!]
\caption{In-distribution Datasets Information}
\centering
\begin{tabular}{cccc}
\toprule
Datasets & \#Images & \#Classes & Target Domain\\
\midrule
CIFAR10-C & 10,000 & 10 & Corruption\\
CIFAR100-C & 10,000 & 100 & Corruption\\
ImageNet-C & 50,000 & 1,000 & Corruption\\
ImageNet-R & 30,000 & 200 & Style Transfer\\
VisDA-C & 55,388 & 12 & Style Transfer\\
\bottomrule
\end{tabular}
\label{tab:id_datasets}
\end{table}
\subsection{Open-world Benchmarks}
The open-world benchmarks employed in our experiments include both In-Distribution (ID) and Out-Of-Distribution (OOD) samples. Specifically, the ID samples are drawn from two types of target domains, i.e., corruption and style transfer. For corruption domain, \textbf{CIFAR10-C}~\cite{DBLP:conf/iclr/HendrycksD19}, \textbf{CIFAR100-C}~\cite{DBLP:conf/iclr/HendrycksD19},  \textbf{ImageNet-C}~\cite{DBLP:conf/iclr/HendrycksD19} are introduced, while for style transfer domain, we have \textbf{ImageNet-R}~\cite{DBLP:conf/iccv/HendrycksBMKWDD21} and \textbf{VisDA-C}~\cite{DBLP:journals/corr/abs-1710-06924}. Details of the ID samples are presented in Table~\ref{tab:id_datasets}.
Besides, we select Gaussian noise and five real-world datasets with distributions different from the ID samples to serve as the OOD datasets: \textbf{Noise}, composed with random Gaussian noise, \textbf{MNIST}~\cite{DBLP:journals/spm/Deng12}, \textbf{SVHN}~\cite{netzer2011reading}, \textbf{Tiny-ImageNet}~\cite{le2015tiny}, {\textbf{CIFAR100-C} and \textbf{CIFAR10-C}. }

We concatenate ID and OOD samples at an OOD sample ratio of $\frac{|\mathcal{S}_{T_O}|}{|\mathcal{S}_{T_I}|}=1$ to form the open-world benchmarks. Specifically, for CIFAR10-C and CIFAR100-C, the OOD samples include Noise, MNIST, Tiny-ImageNet, and {CIFAR100-C/CIFAR10-C}. For ImageNet-C, ImageNet-R, and VisDA-C, the OOD samples are Noise, MNIST, and SVHN.

    



\subsection{Evaluation metrics}
To evaluate the OWTTA performance, similar to~\cite{DBLP:conf/iccv/Li0SJ23}, we introduce three evaluation metrics:
\begin{equation*}
\begin{aligned}
    ACC_I =& \frac{\sum_{\mathbf{x}_i \in \mathcal{D}_{T_I}} \mathbf{1}\left(\hat{y}_i=\Psi\left(\mathbf{x}_i \right)\right) \cdot \mathbf{1}(\Psi(\mathbf{x}_i) \in \mathcal{Y}_s)}
    {\sum_{\mathbf{x}_i \in \mathcal{D}_{T_I}} \mathbf{1}(\Psi(\mathbf{x}_i) \in \mathcal{Y}_s)},\\
    ACC_O =& \frac{\sum_{\mathbf{x}_i\in \mathcal{D}_{T_O}} \mathbf{1}(\hat{y}_i \in \mathcal{Y}_O) \cdot \mathbf{1}(\Psi(\mathbf{x}_i) \in \mathcal{Y}_O)}
    {\sum_{\mathbf{x}_i \in \mathcal{D}_{T_O}} \mathbf{1}(\Psi(\mathbf{x}_i) \in \mathcal{Y}_O) },\\
    ACC_H =&~2 \cdot  \frac{ACC_I \cdot ACC_O}{ACC_I + ACC_O},
\end{aligned} 
\end{equation*}
where 
$ACC_I$ and $ACC_O$ evaluate the classification accuracy of ID and OOD samples, respectively. $ACC_H$ is the harmonic mean of $ACC_I$ and $ACC_O$ measuring the overall OWTTA performance.


\subsection{Baselines}
The baselines in our experiments are summarized as follows.
\textbf{TEST}: The classification results are obtained directly from the pre-trained model without any further adaptation, serving as a baseline for comparison. \textbf{BN}~\cite{DBLP:conf/icml/IoffeS15}: The batch normalization layers of the pre-trained network are unfrozen, allowing them to adjust their parameters based on the test samples. 
\textbf{TENT}~\cite{DBLP:conf/iclr/WangSLOD21}: It introduces two affine parameters to adjust the batch normalization layers of the model and updates these parameters by minimizing the entropy loss of target samples. 
\textbf{SHOT}~\cite{DBLP:conf/icml/LiangHF20}: It keeps the classifier parameters fixed while using the pre-trained feature encoder as the initialization for learning in the target domain.
\textbf{OSTTA}~\cite{DBLP:conf/iccv/LeeDCC23}: This method filters out low-confidence OOD samples and adapts the model by minimizing the entropy of high-confidence samples.
{\textbf{EATA}~\cite{DBLP:conf/icml/NiuW0CZZT22}: This method actively selects high-confidence test samples for adaptation while filtering out low-confidence ones, and mitigates catastrophic forgetting through Fisher regularization.}
{\textbf{RMT}~\cite{DBLP:conf/cvpr/DoblerM023}: It  uses symmetrical cross-entropy as a consistency loss and leverages contrastive learning for feature alignment, reducing error accumulation in continual and gradual test-time adaptation.
\textbf{CoTTA}\cite{DBLP:conf/cvpr/0013FGD22}: This method mitigates error accumulation by generating weight-averaged and augmentation-averaged pseudo-labels and alleviates catastrophic forgetting via stochastic restoration of source pre-trained weights during continual test-time adaptation.}
\textbf{UniEnt}~\cite{gao2024unified}: It uses a two-component GMMs to partition ID and OOD samples based on similarity scores between features and prototypes. The entropy of ID samples is minimized, while that of OOD samples is maximized for better adaptation.
\textbf{OWT3}~\cite{DBLP:conf/iccv/Li0SJ23}: This method adapts the model to the target domain by aligning it with statistical information from the source domain and generating new prototypes for OOD samples. However, in OWTTA, since source statistics are unavailable, we replace the prototypes in OWT3 with normalized classifier weights and remove the alignment component of the loss function.

For the baselines not tailored for OWTTA, we follow the approach in~\cite{DBLP:conf/iccv/Li0SJ23} for comparison. Specifically, samples from each batch are first partitioned into the ID and OOD samples using $\tau^*_t$ calculated by Eq.~\eqref{eq:ood}, with the prototypes extracted by pre-trained classifier weights. Then, these methods are applied to the ID samples for the adaptation. 
Following~\cite{gao2024unified}, we report results for OSTTA and UniEnt combined with TENT.
Besides, our evaluation metric differs from those used in OSTTA and UniEnt~\footnote{We ensure the performance of OSTTA and UniEnt using the original metrics remains consistent to their respective papers.}, since we apply the stricter evaluation metrics, identical to those used in OWT3~\cite{DBLP:conf/iccv/Li0SJ23}.

\subsection{Overall Comparisons in the OWTTA Scenario}
In this section, we validate the effectiveness of the proposed method in the OWTTA scenario. The performance is evaluated using $ACC_I$, $ACC_O$ and $ACC_H$. We compare the proposed method with {ten} baselines across all the open-world benchmarks, including both corrupted and style transfer domains with OOD samples. Tables~\ref{tab:cifar10c}, \ref{tab:cifar100c} and \ref{tab:ImagenetC} present the results for the corrupted domains. The results for the style transfer domain are shown in Tables~\ref{tab:imagenet-R} and~\ref{tab:VisDA-C}. These experimental results reveal several noteworthy observations.

1)~Our ReNC consistently achieves superior $ACC_H$ across all baselines, validating its capability to effectively adapt to diverse datasets in the open-world scenarios. {Specifically, on the ImageNet-C dataset, where existing methods struggle to detect OOD samples using fixed prototypes, ReNC achieves consistently higher $ACC_H$, highlighting the effectiveness of our prototype updating mechanism. Meanwhile, on ImageNet-R and VisDA-C datasets, where all methods have relatively high $ACC_O$, our method still improves ID classification accuracy by approximating the neural collapse state in the target domain.}

2)~UniEnt and OSTTA underperform compared to TENT on nearly all the open-world benchmarks, suggesting that GMMs and confidence-based partitioning mechanisms may not align well with their entropy-based optimization goals. In contrast, our ReNC performs data partitioning and prototype updating within the same latent space, making it more robust since both processes are tightly coupled.

3)~The classification accuracy decreases for all methods as the number of classes increases, especially for ImageNet-C, which contains 1000 classes. 
Our results in Table~\ref{tab:ImagenetC} indicate an improvement in $ACC_H$ from 21.37\% to 31.02\%. This improvement can be attributed to the neural collapse approximation mechanism of ReNC. For the ImageNet-C dataset, the number of classes exceeds the batch size, resulting in imbalanced or even missing classes in each batch. While ReNC updates the prototypes stochastically, it effectively addresses this issue and thus achieves better performance compared to other methods.




In conclusion, our ReNC stochastically updates the prototypes to fit the target data through neural collapse approximation, operating without any source domain information. It achieves the best performance across all the open-world benchmarks.


\begin{table*}[htpb!]
\centering
\caption{Open-world test-time adaptation results on CIFAR10-C with OOD samples.
Best marked in bold, second best underlined.}
\setlength{\tabcolsep}{3pt}
\resizebox{\textwidth}{!}{
\begin{tabular}{c cc>{\columncolor{gray!30}}c |cc>{\columncolor{gray!30}}c| cc>{\columncolor{gray!30}}c| cc>{\columncolor{gray!30}}c| cc>{\columncolor{gray!30}}c}
\toprule[1pt]
\multirow{2}{*}{CIFAR10-C} & \multicolumn{3}{c}{Noise} & \multicolumn{3}{c}{MNIST} & \multicolumn{3}{c}{SVHN} & \multicolumn{3}{c}{Tiny} & \multicolumn{3}{c}{{CIFAR100-C}} \\ 
\cmidrule{2-16} 

& $ACC_I$  & $ACC_O$ & $ACC_H$ & $ACC_I$  & $ACC_O$ & $ACC_H$ & $ACC_I$  & $ACC_O$ & $ACC_H$ & $ACC_I$  & $ACC_O$ & $ACC_H$ & $ACC_I$  & $ACC_O$ & $ACC_H$   \\ 
\cmidrule{2-16}

TEST & 69.19 & \underline{99.96} & 81.78 & 63.61 & 91.11 & 74.92 & 63.54 & 90.91 & 74.80 & 61.09 & 77.48 & 68.32 & 61.27 & \textbf{76.58} & 68.07 \\
BN & 39.44 & 60.57 & 47.77 & 60.19 & 69.56 & 64.54 & 69.21 & 80.17 & 74.29 & 69.02 & 75.15 & 71.95 & 68.83 & 73.33 & 71.01 \\
TENT & 42.75 & 61.99 & 50.60 & 63.21 & 72.80 & 67.67 & 70.84 & 82.92 & 76.41 & 69.70 & 76.23 & 72.82 & 69.22 & 74.07 & 71.56 \\
SHOT & 42.71 & 60.72 & 50.15 & \underline{64.57} & 73.51 & 68.75 & 71.12 & 84.37 & 77.18 & 69.51 & 77.06 & \underline{73.09} & 69.10 & 74.76 & \underline{71.82} \\
OSTTA & 53.43 & 13.69 & 21.80 & 53.88 & 36.94 & 43.83 & 62.76 & 41.20 & 49.74 & \textbf{74.85} & 32.38 & 45.20 & \textbf{76.39} & 29.80 & 42.87 \\
{EATA} & {40.65} & {64.14} & {49.76} & {58.04} & {72.63} & {64.52} & {66.62} & {82.29} & {73.63} & {66.72} & {77.37} & {71.65} & {66.83} & {75.00} & {70.68} \\
{RMT} & {45.50} & {44.32} & {44.90} & {73.35} & {\textbf{99.66}} & {\underline{84.50}} & {\underline{71.97}} & {88.65} & {\underline{79.44}} & {68.65} & {\textbf{78.11}} & {73.08} & {66.84} & {74.51} & {70.47} \\
{CoTTA} & {35.89} & {68.20} & {47.03} & {60.20} & {69.25} & {64.41} & {69.20} & {80.11} & {74.26} & {69.07} & {75.14} & {71.98} & {68.67} & {73.65} & {71.07} \\
UniEnt & 39.29 & 59.97 & 47.48 & 59.70 & 69.79 & 64.35 & 68.47 & 80.33 & 73.93 & 68.58 & 75.46 & 71.86 & 68.80 & 73.50 & 71.07 \\
OWT3 & \underline{69.41} & \underline{99.96} & \underline{81.93} & 63.89 & 91.10 & 75.11 & 63.72 & \underline{90.97} & 74.94 & 61.05 & 77.56 & 68.32 & 61.37 & \underline{76.47} & 68.09 \\
Ours & \textbf{79.14} & \textbf{99.98} & \textbf{88.35} & \textbf{82.23} & \underline{98.12} & \textbf{89.47} & \textbf{77.00} & \textbf{91.54} & \textbf{83.64} & \underline{70.76} & \underline{77.63} & \textbf{74.04} & \underline{70.26} & 75.06 & \textbf{72.58}
\\ 
\bottomrule[1pt]
\end{tabular}}
\label{tab:cifar10c}
\end{table*}

\begin{table*}[htpb!]
\centering
\caption{Open-world test-time adaptation results on CIFAR100-C with OOD samples. Best marked in bold, second best underlined.}
\setlength{\tabcolsep}{3pt}
\resizebox{\textwidth}{!}{
\begin{tabular}{c cc>{\columncolor{gray!30}}c| cc>{\columncolor{gray!30}}c| cc>{\columncolor{gray!30}}c| cc>{\columncolor{gray!30}}c| cc>{\columncolor{gray!30}}c}
\toprule[1pt]
\multirow{2}{*}{CIFAR100-C} & \multicolumn{3}{c}{Noise} & \multicolumn{3}{c}{MNIST} & \multicolumn{3}{c}{SVHN} & \multicolumn{3}{c}{Tiny} & \multicolumn{3}{c}{{CIFAR10-C}} \\ 
\cmidrule{2-16} 
& $ACC_I$  & $ACC_O$ & $ACC_H$ & $ACC_I$  & $ACC_O$ & $ACC_H$ & $ACC_I$  & $ACC_O$ & $ACC_H$ & $ACC_I$  & $ACC_O$ & $ACC_H$ & $ACC_I$  & $ACC_O$ & {$ACC_H$}   \\ 
\cmidrule{2-16}

TEST & 25.30 & \textbf{99.99} & 40.38 & 24.76 & 46.81 & 32.39 & 26.34 & 91.65 & 40.92 & 26.20 & 86.68 & 40.24 & 25.79 & 87.86 & 39.88 \\
BN & 21.22 & 93.29 & 34.58 & 26.09 & \underline{93.21} & 40.77 & 30.60 & \underline{95.42} & 46.34 & 31.41 & 89.69 & \underline{46.53} & 31.95 & 89.93 & 47.15 \\
TENT & 24.45 & 95.75 & 38.95 & 27.50 & 91.49 & 42.29 & 32.05 & 95.17 & 47.95 & 31.33 & 89.64 & 46.43 & 32.08 & \underline{89.93} & 47.29 \\
SHOT & 25.98 & 96.62 & 40.95 & \underline{28.26} & 87.36 & \underline{42.71} & 32.42 & 95.14 & \underline{48.36} & 31.18 & 89.71 & 46.28 & 32.07 & \textbf{90.10} & 47.30 \\
OSTTA & 12.20 & 40.58 & 18.76 & 17.97 & 46.56 & 25.93 & \underline{33.33} & 62.59 & 43.50 & \textbf{36.81} & 52.15 & 43.16 & \textbf{42.75} & 43.67 & 43.21 \\
{EATA} & {22.73} & {94.06} & {36.61} & {25.58} & {92.08} & {40.04} & {29.57} & {\textbf{96.20}} & {45.24} & {30.35} & {\textbf{90.03}} & {45.40} & {31.44} & {89.79} & {46.57} \\
{RMT} & {\underline{27.68}} & {\underline{99.92}} & {\underline{43.35}} & {26.40} & {83.61} & {40.13} & {27.92} & {89.52} & {42.56} & {27.58} & {81.59} & {41.22} & {28.34} & {84.31} & {42.42} \\
{CoTTA} & {20.93} & {93.25} & {34.19} & {25.23} & {92.85} & {39.68} & {30.09} & {95.26} & {45.73} & {31.17} & {\underline{89.85}} & {46.28} & {31.86} & {89.90} & {47.05} \\
UniEnt & 20.05 & 94.48 & 33.08 & 25.54 & 92.86 & 40.06 & 31.36 & 94.48 & 47.09 & 31.20 & 89.19 & 46.23 & 32.52 & 89.03 & \underline{47.64} \\
OWT3 & 25.54 & \textbf{99.99} & 40.69 & 24.51 & 45.51 & 31.86 & 26.48 & 91.70 & 41.09 & 26.34 & 86.68 & 40.40 & 25.93 & 87.82 & 40.04 \\
Ours & \textbf{48.87} & 97.74 & \textbf{65.16} & \textbf{38.75} & \textbf{94.31} & \textbf{54.93} & \textbf{40.96} & 78.22 & \textbf{53.77} & \underline{36.27} & 71.91 & \textbf{48.22} & \underline{35.57} & 76.75 & \textbf{48.61}

\\ 
\bottomrule[1pt]
\end{tabular}}
\label{tab:cifar100c}
\end{table*}

\begin{table}[htpb!]
\centering
\caption{Open-world test-time adaptation results on ImageNet-C with OOD samples. Best marked in bold, second best underlined.}
\setlength{\tabcolsep}{2pt}
\resizebox{0.5\columnwidth}{!}{
\begin{tabular}{c cc>{\columncolor{gray!30}}c |cc>{\columncolor{gray!30}}c| cc>{\columncolor{gray!30}}c}
\toprule[1pt]
\multirow{2}{*}{ImageNet-C} & \multicolumn{3}{c}{Noise} & \multicolumn{3}{c}{MNIST} & \multicolumn{3}{c}{SVHN} \\ 
\cmidrule{2-10} 
& $ACC_I$  & $ACC_O$ & $ACC_H$ & $ACC_I$  & $ACC_O$ & $ACC_H$ & $ACC_I$  & $ACC_O$ & $ACC_H$   \\ 
\cmidrule{2-10}
TEST & 11.79 & 40.85 & 18.30 & 11.33 & 59.57 & 19.04 & 11.62 & 82.04 & 20.36 \\
BN & 9.73 & 66.50 & 16.98 & 15.20 & 74.24 & 25.23 & 16.24 & 77.93 & 26.88 \\
TENT & 9.31 & 68.49 & 16.39 & 15.26 & 72.67 & 25.22 & 17.28 & 78.79 & 28.34 \\
SHOT & 9.70 & 64.57 & 16.87 & 14.76 & 71.69 & 24.48 & 17.70 & 79.63 & 28.96 \\
OSTTA & 13.90 & 65.34 & 22.92 & 20.54 & 57.34 & 30.25 & \underline{21.93} & 59.43 & 32.04 \\
{EATA} & {15.92} & {88.86} & {27.00} & {\underline{20.72}} & {\underline{93.21}} & {\underline{33.90}} & {20.69} & {93.14} & {\underline{33.86}} \\
{RMT} & {\underline{30.38}} & {\underline{99.76}} & {\underline{46.58}} & {11.92} & {90.93} & {21.08} & {12.30} & {\underline{96.85}} & {21.83} \\
{CoTTA} & {8.75} & {67.48} & {15.49} & {13.78} & {82.82} & {23.63} & {14.13} & {74.66} & {23.76} \\
UniEnt & 8.29 & 73.44 & 14.90 & 14.13 & 79.71 & 24.00 & 14.81 & 82.88 & 25.13 \\
OWT3 & 9.76 & 68.03 & 17.07 & 15.41 & 75.29 & 25.58 & 16.45 & 78.67 & 27.21 \\
Ours & \textbf{36.96} & \textbf{99.78} & \textbf{53.94} & \textbf{35.61} & \textbf{98.79} & \textbf{52.35} & \textbf{36.73} & \textbf{97.87} & \textbf{53.41}
\\
\bottomrule[1pt]
\end{tabular}}
\label{tab:ImagenetC}
\end{table}

\begin{table}[htpb!]
\centering
\caption{Open-world test-time adaptation results on ImageNet-R with OOD samples. Best marked in bold, second best underlined.}
\setlength{\tabcolsep}{2pt}
\resizebox{0.5\columnwidth}{!}{
\begin{tabular}{c cc>{\columncolor{gray!30}}c| cc>{\columncolor{gray!30}}c |cc>{\columncolor{gray!30}}c}
\toprule[1pt]
\multirow{2}{*}{ImageNet-R} & \multicolumn{3}{c}{Noise} & \multicolumn{3}{c}{MNIST} & \multicolumn{3}{c}{SVHN} \\ 
\cmidrule{2-10} 
& $ACC_I$  & $ACC_O$ & $ACC_H$ & $ACC_I$  & $ACC_O$ & $ACC_H$ & $ACC_I$  & $ACC_O$ & $ACC_H$   \\ 
\cmidrule{2-10} 
TEST & 24.71 & \textbf{100.00} & \underline{39.63} & \underline{23.60} & \textbf{99.81} & \underline{38.17} & 22.46 & \textbf{98.89} & \underline{36.61} \\
BN & 16.38 & 89.76 & 27.70 & 18.70 & 82.97 & 30.52 & 19.03 & 86.08 & 31.17 \\
TENT & 16.87 & 93.23 & 28.57 & 19.81 & 84.72 & 32.11 & 20.09 & 87.99 & 32.71 \\
SHOT & 17.27 & 96.24 & 29.28 & 19.70 & 81.43 & 31.72 & 20.44 & 89.61 & 33.29 \\
OSTTA & \underline{25.46} & 58.59 & 35.50 & 22.46 & 49.67 & 30.93 & \underline{25.66} & 58.02 & 35.58 \\
{EATA} & {16.21} & {91.71} & {27.55} & {18.95} & {86.24} & {31.07} & {18.95} & {88.37} & {31.21} \\
{RMT} & {8.99} & {73.27} & {16.02} & {12.27} & {93.79} & {21.70} & {16.36} & {77.63} & {27.02} \\
{CoTTA} & {15.58} & {85.41} & {26.35} & {17.40} & {86.71} & {28.98} & {17.51} & {83.04} & {28.92} \\
UniEnt & 10.56 & 72.46 & 18.43 & 15.01 & 79.59 & 25.26 & 15.84 & 83.72 & 26.64 \\
OWT3 & 16.56 & 92.13 & 28.07 & 19.13 & 83.75 & 31.15 & 19.42 & 86.93 & 31.75 \\
Ours & \textbf{39.85} & \underline{98.67} & \textbf{56.77} & \textbf{37.99} & \underline{97.32} & \textbf{54.65} & \textbf{39.36} & \underline{95.68} & \textbf{55.78}
\\
\bottomrule[1pt]
\end{tabular}}
\label{tab:imagenet-R}
\end{table}

\begin{table}[htpb!]
\centering
\caption{Open-world test-time adaptation results on VisDA-C with OOD samples. Best marked in bold, second best underlined.}
\setlength{\tabcolsep}{2pt}
\resizebox{0.5\columnwidth}{!}
{
\begin{tabular}{c cc>{\columncolor{gray!30}}c| cc>{\columncolor{gray!30}}c |cc>{\columncolor{gray!30}}c}
\toprule[1pt]
\multirow{2}{*}{VisDA-C} & \multicolumn{3}{c}{Noise} & \multicolumn{3}{c}{MNIST} & \multicolumn{3}{c}{SVHN} \\ 
\cmidrule{2-10} 
& $ACC_I$  & $ACC_O$ & $ACC_H$ & $ACC_I$  & $ACC_O$ & $ACC_H$ & $ACC_I$  & $ACC_O$ & $ACC_H$   \\ 
\cmidrule{2-10} 
TEST & 41.27 & \textbf{100.00} & 58.43 & 43.12 & \underline{97.41} & \underline{59.78} & 42.06 & \textbf{99.46} & 59.12 \\
BN & 45.48 & 99.64 & 62.45 & 42.06 & 79.96 & 55.12 & 42.80 & 88.77 & 57.75 \\
TENT & 51.57 & \textbf{100.00} & 68.05 & 41.50 & 79.96 & 54.64 & 44.57 & 88.40 & \underline{59.26} \\
SHOT & 42.39 & 99.94 & 59.53 & 42.69 & 68.08 & 52.48 & 43.45 & 72.69 & 54.33 \\
OSTTA & 39.60 & 54.33 & 45.81 & \underline{46.41} & 29.81 & 36.30 & \underline{44.91} & 45.55 & 45.23 \\
{EATA} & {43.12} & {99.68} & {60.20} & {41.49} & {81.13} & {54.90} & {41.75} & {89.25} & {56.89} \\
{RMT} & {25.87} & {74.63} & {38.42} & {26.09} & {99.98} & {41.38} & {30.36} & {63.64} & {41.11} \\
{CoTTA} & {44.01} & {99.56} & {61.04} & {40.71} & {83.91} & {54.82} & {40.89} & {89.86} & {56.20} \\
UniEnt & \underline{54.84} & 95.99 & \underline{69.80} & 41.92 & 80.13 & 55.04 & 42.83 & 88.53 & 57.73 \\
OWT3 & 46.12 & \underline{99.99} & 63.12 & 42.14 & 77.39 & 54.57 & 43.04 & 91.60 & 58.56 \\
Ours & \textbf{55.96} & 99.63 & \textbf{71.67} & \textbf{47.42} & \textbf{99.17} & \textbf{64.16} & \textbf{53.93} & \underline{98.23} & \textbf{69.63}
\\
\bottomrule[1pt]
\end{tabular}}
\label{tab:VisDA-C}
\end{table}

\newpage

\subsection{Connections between TTA and Neural Collapse}\label{sec:NC_}
In this section, we investigate the connection between TTA and neural collapse to validate that the proposed method aligns with the intrinsic properties of well-trained neural networks.

\textbf{Neural collapse in the source domain:}
Assumption~\ref{assum:well-train} states that the pre-trained model has already collapsed in the source domain. To validate this, we calculate the $\mathcal{NC}_1$ and $\mathcal{NC}_3$ of the pre-trained model using its training data. $\mathcal{NC}_1$ reflects how well the features collapse to their class-means and $\mathcal{NC}_3$ measures the self-duality between class-mean and the classifier weights. The lower the values of these two metrics, the closer the model is to achieving the neural collapse state.
As shown in Table~\ref{tab:nc_original}, the models trained on different datasets have indeed collapsed on their respective datasets as indicated by low $\mathcal{NC}_1$ and $\mathcal{NC}_3$ values\footnote{Due to hardware limitations, we assess the neural collapse state of the ImageNet dataset using its validation set.}.

\begin{table}[htpb!] 
\centering
\caption{$\mathcal{NC}_1$ and $\mathcal{NC}_3$ on each training dataset. Lower values indicate closer to the neural collapse state.}
\setlength{\tabcolsep}{4.5pt}
\resizebox{0.5\columnwidth}{!}{
\begin{tabular}{cccccccc}
\toprule[1pt]
\multicolumn{2}{c}{CIFAR10} & \multicolumn{2}{c}{CIFAR100} & \multicolumn{2}{c}{ImageNet} & \multicolumn{2}{c}{VisDA}\\
\cmidrule{1-8}
 $\mathcal{NC}_1$ & $\mathcal{NC}_3$
& $\mathcal{NC}_1$ & $\mathcal{NC}_3$
& $\mathcal{NC}_1$ & $\mathcal{NC}_3$
& $\mathcal{NC}_1$ & $\mathcal{NC}_3$
\\ 
\cmidrule(lr){1-2} \cmidrule(lr){3-4} \cmidrule(lr){5-6} \cmidrule(lr){7-8}
0.1712 & 0.2466 & 0.7541 & 0.6235 & 2.3275 & 1.1968 & 0.3031 & 0.5146\\
\bottomrule[1pt]
\end{tabular}}
\label{tab:nc_original}
\end{table}

\begin{table}[htpb!] 

\centering
\caption{{The ratio $\mathbf{b}$ to $\langle\mathbf{w},\mathbf{c}\rangle$ quantifies their relative influence. Lower values imply that the bias term is increasingly negligible in the neural collapse state.}}
\setlength{\tabcolsep}{3pt}
\resizebox{0.5\columnwidth}{!}{
\begin{tabular}{cccccc}
\toprule[1pt]
Dataset&\multicolumn{1}{c}{CIFAR10} & \multicolumn{1}{c}{CIFAR100} & \multicolumn{1}{c}{ImageNet} & \multicolumn{1}{c}{VisDA}\\
\cmidrule{1-5}
Ratio& $1.526\times10^{-2}$ & $7.693\times10^{-3}$ & $1.116\times10^{-3}$ & $1.873\times10^{-3}$
\\ 
\bottomrule[1pt]
\end{tabular}}
\label{tab:assum2 ratio}
\end{table}

\begin{table*}[htpb!]
\caption{$\mathcal{NC}_1$ and $\mathcal{NC}_3$ on CIFAR10-C and CIFAR100-C with OOD samples. Best results are marked in bold, second best are underlined. The gray background indicates the highest $ACC_S$.}
\centering
\resizebox{0.8\columnwidth}{!}{
\begin{tabular}{c cc| cc| cc| cc| cc}
\toprule[1pt]

\multirow{2}{*}{CIFAR10-C} & \multicolumn{2}{c}{Noise} & \multicolumn{2}{c}{MNIST} & \multicolumn{2}{c}{SVHN} & \multicolumn{2}{c}{Tiny} & \multicolumn{2}{c}{{CIFAR100-C}} \\
\cmidrule{2-11}
 & $\mathcal{NC}_1$ & $\mathcal{NC}_3$ & $\mathcal{NC}_1$ & $\mathcal{NC}_3$ & $\mathcal{NC}_1$ & $\mathcal{NC}_3$ & $\mathcal{NC}_1$ & $\mathcal{NC}_3$ & $\mathcal{NC}_1$ & $\mathcal{NC}_3$ \\
 \cmidrule{2-11}

TEST & 0.7386 & 0.3548 & 0.7386 & 0.3548 & 0.7386 & 0.3548 & 0.7386 & 0.3548 & 0.7386 & 0.3548 \\
BN & 1.5176 & 0.4034 & 0.8026 & 0.3637 & 0.6209 & 0.3189 & 0.6052 & 0.3204 & 0.5970 & 0.3193 \\
TENT & 1.4132 & 0.3917 & 0.7314 & 0.3497 & 0.5924 & \underline{0.3151} & 0.5946 & \textbf{0.3179} & 0.5884 & \textbf{0.3179} \\
SHOT & 1.3871 & 0.3914 & 0.7057 & \underline{0.3461} & \underline{0.5815} & \textbf{0.3147} & \underline{0.5914} & \underline{0.3180} & 0.5863 & \underline{0.3185} \\
OSTTA & 1.4157 & 0.4056 & 0.7922 & 0.3633 & 0.6154 & 0.3194 & \cellcolor{gray!30}0.5944 & \cellcolor{gray!30}0.3212 & \cellcolor{gray!30}\underline{0.5852} & \cellcolor{gray!30}0.3203 \\
{EATA} & {1.5056} & {0.4046} & {0.8125} & {0.3665} & {0.6307} & {0.3205} & {0.6157} & {0.3224} & {0.6065} & {0.3215} \\
{RMT} & {1.3723} & {0.4611} & {\underline{0.6609}} & {0.3963} & {0.6399} & {0.3979} & {0.6970} & {0.4060} & {0.6858} & {0.4143} \\
{CoTTA} & {1.9769} & {0.4753} & {0.7999} & {0.3626} & {0.6208} & {0.3184} & {0.6074} & {0.3206} & {0.5963} & {0.3189} \\
UniEnt & 1.4784 & 0.4017 & 0.8185 & 0.3666 & 0.6270 & 0.3195 & 0.6101 & 0.3201 & 0.6039 & 0.3200 \\
OWT3 & \underline{0.7358} & \underline{0.3545} & 0.7358 & 0.3545 & 0.7350 & 0.3545 & 0.7366 & 0.3545 & 0.7368 & 0.3545 \\
Ours & \cellcolor{gray!30}\textbf{0.5731} & \cellcolor{gray!30}\textbf{0.3202} & \cellcolor{gray!30}\textbf{0.5782} & \cellcolor{gray!30}\textbf{0.3174} & \cellcolor{gray!30}\textbf{0.5667} & \cellcolor{gray!30}0.3203 & \textbf{0.5795} & 0.3195 & \textbf{0.5761} & 0.3202
\\
\cmidrule{1-11}
\toprule[1pt]
\multirow{2}{*}{CIFAR100-C} & \multicolumn{2}{c}{Noise} & \multicolumn{2}{c}{MNIST} & \multicolumn{2}{c}{SVHN} & \multicolumn{2}{c}{Tiny} & \multicolumn{2}{c}{{CIFAR10-C}} \\
\cmidrule{2-11}
  & $\mathcal{NC}_1$ & $\mathcal{NC}_3$ & $\mathcal{NC}_1$ & $\mathcal{NC}_3$ & $\mathcal{NC}_1$ & $\mathcal{NC}_3$ & $\mathcal{NC}_1$ & $\mathcal{NC}_3$ & $\mathcal{NC}_1$ & $\mathcal{NC}_3$ \\
\cmidrule{2-11}
TEST & 9.5278 & 0.8884 & 9.5278 & 0.8884 & 9.5278 & 0.8884 & 9.5278 & 0.8884 & 9.5278 & 0.8884 \\
BN & 14.1927 & 0.9529 & 10.6814 & 0.9139 & 8.2031 & 0.8764 & 7.8558 & 0.8689 & 7.7276 & 0.8709 \\
TENT & 11.6919 & 0.9177 & 9.4333 & 0.8909 & \underline{7.7855} & 0.8649 & 7.7511 & 0.8631 & 7.6456 & 0.8652 \\
SHOT & 10.5667 & 0.9027 & 8.8852 & \underline{0.8815} & \textbf{7.6591} & \underline{0.8608} & 7.7561 & \underline{0.8610} & 7.6657 & \underline{0.8631} \\
OSTTA & 13.3452 & 0.9586 & 10.3678 & 0.9193 & 7.9534 & 0.8755 & \cellcolor{gray!30}\textbf{7.4570} & \cellcolor{gray!30}0.8673 & \cellcolor{gray!30}\textbf{6.9739} & \cellcolor{gray!30}0.8634 \\
{EATA} & {13.0616} & {0.9414} & {10.9230} & {0.9124} & {7.9622} & {0.8725} & {\underline{7.6159}} & {0.8682} & {\underline{7.3165}} & {0.8675} \\
{RMT} & {\underline{8.1638}} & {0.9008} & {\underline{8.2756}} & {0.9038} & {8.2989} & {0.9036} & {8.5444} & {0.9172} & {8.0033} & {0.9041} \\
{CoTTA} & {14.6210} & {0.9554} & {11.2101} & {0.9200} & {8.4420} & {0.8792} & {7.9485} & {0.8704} & {7.8608} & {0.8727} \\
UniEnt & 14.2197 & 0.9363 & 11.7456 & 0.9174 & 8.3867 & 0.8729 & 8.2405 & 0.8713 & 8.1938 & 0.8757 \\
OWT3 & 9.4840 & \underline{0.8877} & 9.5121 & 0.8882 & 9.4805 & 0.8876 & 9.4835 & 0.8877 & 9.4799 & 0.8876 \\
Ours & \cellcolor{gray!30}\textbf{7.7409} & \cellcolor{gray!30}\textbf{0.8606} & \cellcolor{gray!30}\textbf{7.6476} & \cellcolor{gray!30}\textbf{0.8700} & \cellcolor{gray!30}8.0640 & \cellcolor{gray!30}\textbf{0.8604} & 7.7153 & \textbf{0.8557} & 7.6808 & \textbf{0.8577}\\
\bottomrule[1pt]
\end{tabular}}
\label{tab:nc_cifar100}
\end{table*}

\begin{table}[htpb!]
\vspace{-0.1in}
\caption{$\mathcal{NC}_1$ and $\mathcal{NC}_3$ on ImageNet-C with OOD samples. Best results marked in bold, second best underlined. The gray background indicates the highest $ACC_S$.}
\centering
\setlength{\tabcolsep}{5.5pt}
\resizebox{0.65\columnwidth}{!}{
\begin{tabular}{c cc| cc| cc}
\toprule[1pt]
\multirow{2}{*}{ImageNet-C} & \multicolumn{2}{c}{Noise} & \multicolumn{2}{c}{MNIST} & \multicolumn{2}{c}{SVHN} \\
\cmidrule{2-7}
  & $\mathcal{NC}_1$ & $\mathcal{NC}_3$ & $\mathcal{NC}_1$ & $\mathcal{NC}_3$ & $\mathcal{NC}_1$ & $\mathcal{NC}_3$  \\
 \cmidrule{2-7}
TEST & \underline{3.7824} & 1.2762 & \underline{3.7824} & 1.2762 & 3.7824 & 1.2762 \\
BN & 7.6031 & 1.2733 & 6.0601 & 1.2620 & 6.7761 & 1.2581 \\
TENT & 7.4605 & 1.2755 & 5.6129 & 1.2635 & 6.3444 & 1.2574 \\
SHOT & 7.6685 & 1.2759 & 5.8075 & 1.2610 & 6.3542 & 1.2557 \\
OSTTA & 6.0655 & 1.2586 & 4.1373 & 1.2554 & \underline{4.1407} & 1.2557 \\
{EATA} & {6.9685} & {\underline{1.2501}} & {5.3900} & {\textbf{1.2427}} & {5.4280} & {1.2527} \\
{RMT} & {\textbf{1.9977}} & {1.4037} & {\textbf{2.7490}} & {1.3911} & {\textbf{2.0345}} & {1.4074} \\
{CoTTA} & {7.7565} & {1.2733} & {7.2648} & {1.2652} & {7.9593} & {1.2616} \\
UniEnt & 8.0681 & 1.2695 & 5.6453 & 1.2506 & 5.8017 & \underline{1.2496} \\
OWT3 & 7.4846 & 1.2724 & 5.9790 & 1.2609 & 6.7146 & 1.2571 \\
Ours & \cellcolor{gray!30}5.6142 & \cellcolor{gray!30}\textbf{1.2418} & \cellcolor{gray!30}5.3901 & \cellcolor{gray!30}\underline{1.2436} & \cellcolor{gray!30}5.3158 & \cellcolor{gray!30}\textbf{1.2434}\\
\bottomrule[1pt]
\end{tabular}}
\label{tab:nc_imagenet-C}
\end{table}

\begin{table}[htpb!]
\vspace{-0.1in}
\caption{$\mathcal{NC}_1$ and $\mathcal{NC}_3$ on ImageNet-R with OOD samples. Best results marked in bold, second best underlined. The gray background indicates the highest $ACC_S$.}
\centering
\resizebox{0.65\columnwidth}{!}{
\setlength{\tabcolsep}{4.85pt}
\begin{tabular}{c cc| cc| cc}
\toprule[1pt]
\multirow{2}{*}{ImageNet-R} & \multicolumn{2}{c}{Noise} & \multicolumn{2}{c}{MNIST} & \multicolumn{2}{c}{SVHN} \\
\cmidrule{2-7}
  & $\mathcal{NC}_1$ & $\mathcal{NC}_3$ & $\mathcal{NC}_1$ & $\mathcal{NC}_3$ & $\mathcal{NC}_1$ & $\mathcal{NC}_3$  \\
 \cmidrule{2-7}

TEST & \underline{20.6384} & 1.0441 & \underline{20.6384} & \underline{1.0441} & \underline{20.6384} & \underline{1.0441} \\
BN & 30.1058 & 1.0541 & 28.6891 & 1.0577 & 27.8901 & 1.0540 \\
TENT & 27.4597 & 1.0478 & 27.5288 & 1.0549 & 26.6687 & 1.0507 \\
SHOT & 26.1513 & \underline{1.0437} & 27.2900 & 1.0539 & 25.9399 & 1.0480 \\
OSTTA & 25.8711 & 1.0483 & 24.8492 & 1.0584 & 24.3873 & 1.0555 \\
{EATA} & {29.5161} & {1.0521} & {28.0359} & {1.0552} & {27.4065} & {1.0521} \\
{RMT} & {44.8059} & {1.1647} & {37.8196} & {1.1126} & {39.1313} & {1.1318} \\
{CoTTA} & {32.4597} & {1.0595} & {31.2893} & {1.0632} & {30.3840} & {1.0595} \\
UniEnt & 33.4932 & 1.0751 & 26.7478 & 1.0536 & 26.0775 & 1.0478 \\
OWT3 & 28.9696 & 1.0513 & 27.6794 & 1.0548 & 27.1448 & 1.0512 \\
Ours & \cellcolor{gray!30}\textbf{19.7501} & \cellcolor{gray!30}\textbf{1.0345} & \cellcolor{gray!30}\textbf{19.3365} & \cellcolor{gray!30}\textbf{1.0369} & \cellcolor{gray!30}\textbf{19.2924} & \cellcolor{gray!30}\textbf{1.0383}
\\
\bottomrule[1pt]
\end{tabular}}
\label{tab:nc_imagenet-R}
\end{table}

\begin{table}[htbp!]
\vspace{-0.1in}
\caption{$\mathcal{NC}_1$ and $\mathcal{NC}_3$ on VisDA-C with OOD samples. Best results marked in bold, second best underlined. The gray background indicates the highest $ACC_S$.}
\centering
\begin{tabular}{c cc| cc| cc}
\toprule[1pt]
\multirow{2}{*}{VisDA-C} & \multicolumn{2}{c}{Noise} & \multicolumn{2}{c}{MNIST} & \multicolumn{2}{c}{SVHN} \\
\cmidrule{2-7}
  & $\mathcal{NC}_1$ & $\mathcal{NC}_3$ & $\mathcal{NC}_1$ & $\mathcal{NC}_3$ & $\mathcal{NC}_1$ & $\mathcal{NC}_3$  \\
 \cmidrule{2-7}

TEST & 1.8591 & 0.8081 & 1.8591 & 0.8081 & 1.8591 & 0.8081 \\
BN & 1.3001 & 0.7268 & 1.2937 & 0.7217 & 1.2915 & 0.7126 \\
TENT & 1.1664 & 0.7034 & 1.3417 & 0.7287 & \underline{1.1755} & 0.7005 \\
SHOT & \textbf{1.0559} & \textbf{0.6944} & \textbf{1.1516} & 0.7098 & \textbf{1.1388} & 0.7003 \\
OSTTA & 1.4202 & 0.7503 & 1.2309 & 0.748 & 1.2792 & 0.7308 \\
{EATA} & {1.2970} & {0.7272} & {1.2910} & {0.7215} & {1.2934} & {0.7130} \\
{RMT} & {6.2168} & {0.8892} & {6.0925} & {0.8393} & {4.8335} & {0.8780} \\
{CoTTA} & {1.3695} & {0.7298} & {1.3806} & {0.7229} & {1.3659} & {0.7158} \\
UniEnt & \underline{1.1522} & \underline{0.7019} & 1.3653 & \textbf{0.6829} & 1.3198 & \textbf{0.6778} \\
OWT3 & 1.1679 & 0.7083 & \underline{1.1764} & \underline{0.7031} & 1.1955 & \underline{0.6983} \\
Ours & \cellcolor{gray!30}1.1548 & \cellcolor{gray!30}0.7097 & \cellcolor{gray!30}1.3135 & \cellcolor{gray!30}0.7055 & \cellcolor{gray!30}1.2832 & \cellcolor{gray!30}0.7046
\\
\bottomrule[1pt]
\end{tabular}
\label{tab:nc_VisDA-C}
\end{table}

{
\textbf{Empirical validation of Assumption~\ref{assum:neglect}:}
To verify the validity of Assumption~\ref{assum:neglect}, we conduct experiments on a series of classification neural networks.  
According to Theorem~\ref{theo:NC1}, in the context of neural collapse, features $\mathbf{z}$ have collapsed to their class-mean, i.e., prototype $\mathbf{c}$. 
Therefore, to empirically validate this assumption, we measure the ratio of the bias term $\mathbf{b}_j$ to the weight-prototype interaction $\langle\mathbf{w}_{:,j},\mathbf{c}_j\rangle$, i.e., $\frac{1}{K}\sum_j^K\frac{\|\mathbf{b}_j\|}{\langle\mathbf{w}_{:,j},\mathbf{c}_j\rangle}$. As shown in Table~\ref{tab:assum2 ratio}, the weight-prototype interaction significantly outweighs the bias term, {i.e.$\frac{1}{K}\sum_j^K\frac{\|\mathbf{b}_j\|}{\langle\mathbf{w}_{:,j},\mathbf{c}_j\rangle} \ll 1$. This empirical result validates Assumption 2, which posits that the effect of the bias term is negligible in the neural collapse state.}}

\textbf{Neural collapse in the target domain:}
TTA algorithms update model parameters to align with the data distribution of the target domain. Their primary goal is to correct the inaccuracies caused by target samples producing incorrect labels when passed through the pre-trained model. Since the pre-trained model has well-collapsed in the source domain, following Theorem~\ref{theo:NC3} and Theorem~\ref{theo:NC1}, each feature should collapse to its corresponding classifier weight vector. The inaccuracies occur because the features generated by the encoder do not align well with the pre-trained classifier. 
{From this perspective, effective TTA should guide target representations toward a target-domain NC-like geometry.}
To support this claim, we measure the $\mathcal{NC}_1$ and $\mathcal{NC}_3$ on all the open-world benchmarks, and the results are presented in Table{s}~\ref{tab:nc_cifar100}-\ref{tab:nc_VisDA-C}.


{For the corrupted domain, we observe that stronger ID classification performance is often accompanied by better NC-related properties. For example, on CIFAR10-C with Noise and MNIST as OOD samples, ReNC achieves the lowest $\mathcal{NC}_1$ and $\mathcal{NC}_3$ values among all methods, together with the highest $ACC_S$. For CIFAR10-C with Tiny-ImageNet and CIFAR100-C as OOD samples, OSTTA obtains the best $ACC_S$, while ReNC still maintains competitive NC-related metrics. These results suggest that although classification accuracy is not solely determined by NC metrics, preserving compact and well-aligned representations can effectively facilitate the target adaptation process.}

{For the style-transfer domain, as shown in Tables~\ref{tab:nc_imagenet-R} and~\ref{tab:nc_VisDA-C}, ImageNet-R follows a similar trend to the corrupted domain. VisDA-C presents a more challenging case, where SHOT achieves relatively strong collapse metrics but still suffers from poor ID classification performance, even falling below TEST in some settings.} This discrepancy may be due to the distinct training and testing sets in VisDA-C, which cause the source-domain pre-trained classifier to poorly align with the target domain features. 
{ReNC achieves comparable NC-related properties and strong adaptation performance, supporting the effectiveness of neural collapse approximation.}


{In general, neural collapse provides an informative perspective for understanding classification representations. The improved NC-related properties of ReNC offer a plausible explanation for its better performance.}




\subsection{Ablation Study}
In this section, we evaluate the impact of entropy minimization~($\mathcal{L}_e^+$), {the one-hot distribution prior~($\mathcal{L}_d$)}, the alignment of samples with their prototypes~($\mathcal{L}_s$), and finally, the effectiveness of Updating Prototypes~(U.P.). 
The TTA performance is reported using $ACC_H$ on all the open-world benchmarks, as shown in Table~\ref{tab:ablation_cifar10&100} and Table~\ref{tab:ablation imagenet VisDA-C}.
From these tables, we have the following observations:

1)~$\mathcal{L}_s$ may degrade classification performance if the prototypes are not updated. For example, on the CIFAR100-C dataset, applying $\mathcal{L}_s$ leads to a drop in $ACC_H$ with OOD samples such as MNIST, SVHN, Tiny-ImageNet and {CIFAR10-C}. This may be because the prototypes, initialized by the weights of the pre-trained classifier, do not fit well with the target domain. Pushing samples towards these fixed prototypes can mislead the classification process. 
2)~Updating prototypes generally enhances TTA performance on all the {open-world benchmarks}. This can be attributed to the proposed neural collapse approximation mechanism, which updates prototypes stochastically, enabling the prototypes to well align with the target domain. 
3)~Simply minimizing $\mathcal{L}_e^+$ or $\mathcal{L}_d$ can lead to degraded TTA performance. As shown in Table~\ref{tab:ablation imagenet VisDA-C}, when updating prototypes, optimizing the model with $\mathcal{L}_e^+$ or $\mathcal{L}_d$ alone does not provide optimal adaptation results. 
Overall, using any of these losses individually may not yield better adaptation results.


{Our ReNC integrates all the losses and updating mechanisms into a unified pipeline. It encourages variability collapse by pushing samples toward their corresponding prototypes, while preserving balanced embedding structure, which supports more robust and effective adaptation.}

\begin{table}[t!]
\setlength{\tabcolsep}{3pt}
\caption{Ablation study on the CIFAR10-C and CIFAR100-C  with OOD samples. Best results marked in bold, second best underlined.}
\centering
\resizebox{0.65\columnwidth}{!}{
\begin{tabular}{cccc|ccccc}
\toprule[1pt]
\multicolumn{4}{c}{CIFAR10-C} & Noise & MNIST & SVHN & Tiny & {CIFAR100-C} \\
\cmidrule{1-9}
$\mathcal{L}_e^+$ & $\mathcal{L}_d$ & $\mathcal{L}_s$ & U.P. & $ACC_H$ & $ACC_H$ & $ACC_H$ & $ACC_H$ & $ACC_H$ \\
\cmidrule{1-4} \cmidrule{5-9}
\checkmark &  &  &  & 86.05 & 78.60 & 82.10 & 69.25 & 68.04 \\
\checkmark & \checkmark &  &  & 86.07 & 78.56 & 82.06 & 69.27 & 68.00 \\
\checkmark & \checkmark & \checkmark &  & 86.05 & 78.62 & 82.06 & 69.27 & 68.00 \\
\cmidrule(lr){1-9}
& &  \checkmark & \checkmark &88.02	&89.21	&83.22	&\underline{73.77}	& \underline{72.13}
\\
\checkmark &  & \checkmark & \checkmark & \underline{88.07} & \underline{89.24} & 83.30 & 73.42 & 71.47 \\
 & \checkmark & \checkmark & \checkmark & 88.04 & 89.21 & \underline{83.32} & 73.34 & 71.81 \\
\checkmark & \checkmark & \checkmark & \checkmark & \textbf{88.35} & \textbf{89.47} & \textbf{83.64} & \textbf{74.04} & \textbf{72.58} \\
\cmidrule{1-9}
\toprule[1pt]
\multicolumn{4}{c}{CIFAR100-C} & Noise & MNIST & SVHN & Tiny & {CIFAR10-C} \\
\cmidrule{1-9}
$\mathcal{L}_e^+$ & $\mathcal{L}_d$ & $\mathcal{L}_s$ & U.P. & $ACC_H$ & $ACC_H$ & $ACC_H$ & $ACC_H$ & $ACC_H$ \\
\cmidrule{1-9}
\checkmark &  &  &  & 48.99 & 43.49 & 46.21 & 46.22 & 46.51 \\
\checkmark & \checkmark &  &  & 49.28 & 43.50 & 46.29 & 46.43 & 46.61 \\
\checkmark & \checkmark & \checkmark &  & 49.36 & 43.34 & 46.21 & 46.26 & 46.44 \\
\cmidrule(lr){1-9}
 &  & \checkmark & \checkmark &57.06	&\underline{54.58}	&\underline{53.31}	&47.05	&47.46 \\
\checkmark &  & \checkmark & \checkmark & \underline{63.19} & 52.57 & 52.22 & \underline{47.22} & \underline{47.56} \\
 & \checkmark & \checkmark & \checkmark & 54.10 & 51.39 & 50.85 & 46.36 & 46.45 \\
\checkmark & \checkmark & \checkmark & \checkmark & \textbf{65.16} & \textbf{54.93} & \textbf{53.77} & \textbf{48.22} & \textbf{48.61}\\
\bottomrule[1pt]
\end{tabular}}
\label{tab:ablation_cifar10&100}
\vspace{-10pt}
\end{table}

\begin{table}[htpb!]
\centering
\caption{Ablation study on the ImageNet-C, ImageNet-R and VisDA-C with OOD samples. Best results marked in bold, second best underlined.}
\resizebox{0.5\columnwidth}{!}{
\begin{tabular}{cccc|ccc}
\toprule[1pt]
\multicolumn{4}{c}{ImageNet-C} & Noise & MNIST & SVHN \\
\cmidrule{1-7}
$\mathcal{L}_e^+$ & $\mathcal{L}_d$ & $\mathcal{L}_s$ & U.P. & $ACC_H$ & $ACC_H$ & $ACC_H$ \\
\cmidrule{1-7}
\checkmark &  &  &  & 33.11 & 34.96 & 36.56 \\
\checkmark & \checkmark &  &  & 33.25 & 35.00 & 36.47 \\
\checkmark & \checkmark & \checkmark &  & 32.27 & 35.33 & 36.44 \\
\cmidrule(lr){1-7}
& & \checkmark & \checkmark  &\underline{52.84}	& \underline{52.10}	& \underline{52.45}\\
\checkmark &  & \checkmark & \checkmark & 51.67 & 51.23 & 51.53 \\
 & \checkmark & \checkmark & \checkmark & 51.64 & 51.17 & 51.23 \\
\checkmark & \checkmark & \checkmark & \checkmark & \textbf{53.94} & \textbf{52.35} & \textbf{53.41} \\
\cmidrule{1-7}

\toprule[1pt]
\multicolumn{4}{c}{ImageNet-R} & Noise & MNIST & SVHN \\
\cmidrule{1-7}
$\mathcal{L}_e^+$ & $\mathcal{L}_d$ & $\mathcal{L}_s$ & U.P. & $ACC_H$ & $ACC_H$ & $ACC_H$ \\
\cmidrule{1-7}
\checkmark &  &  &  & 42.54 & 39.69 & 36.64 \\
\checkmark & \checkmark &  &  & 42.62 & 39.69 & 36.64 \\
\checkmark & \checkmark & \checkmark &  & 42.74 & 39.66 & 36.58 \\
\cmidrule(lr){1-7}
& & \checkmark & \checkmark  &\underline{56.45}	& \underline{54.16}	& \underline{55.26}\\
\checkmark &  & \checkmark & \checkmark & 55.55 & 52.51 & 54.55 \\
 & \checkmark & \checkmark & \checkmark & 55.54 & 52.49 & 54.59 \\
\checkmark & \checkmark & \checkmark & \checkmark & \textbf{56.77} & \textbf{54.65} & \textbf{55.78} \\
\cmidrule{1-7}

\toprule[1pt]
\multicolumn{4}{c}{VisDA-C} & Noise & MNIST & SVHN \\
\cmidrule{1-7}
$\mathcal{L}_e^+$ & $\mathcal{L}_d$ & $\mathcal{L}_s$ & U.P. & $ACC_H$ & $ACC_H$ & $ACC_H$ \\
\cmidrule{1-7}
\checkmark &  &  &  & 47.07 & 52.81 & 52.94 \\
\checkmark & \checkmark &  &  & 47.08 & 52.83 & 52.90 \\
\checkmark & \checkmark & \checkmark &  & 47.32 & 52.95 & 52.85 \\
\cmidrule(lr){1-7}
& & \checkmark & \checkmark  &\underline{70.39}	& \underline{63.22}	& \underline{68.41}\\
\checkmark &  & \checkmark & \checkmark & 69.81 & 62.68 & 67.86 \\
 & \checkmark & \checkmark & \checkmark & 69.72 & 62.31 & 67.83 \\
\checkmark & \checkmark & \checkmark & \checkmark & \textbf{71.67} & \textbf{64.16} & \textbf{69.63}\\
\bottomrule[1pt]
\end{tabular}}
\label{tab:ablation imagenet VisDA-C}

\end{table}

\newpage

\subsection{{Computational Cost Analysis}}
{In this section, we evaluate the computational cost of each method. Since OOD detection is performed in a batch-wise manner, we set the batch size to 64 and measure the per-sample running time across all benchmarks with Noise OOD samples. The results are reported in Table~\ref{tab:run-time}. 
Under the same experimental settings, TEST and BN show the lowest computational overhead. The proposed ReNC method, while incurring approximately 2–3 times higher cost, achieves significantly better performance. Compared with other OWTTA-specific methods, ReNC demonstrates competitive efficiency relative to UniEnt (0.003270 vs. 0.003955) while achieving superior results. Notably, ReNC runs substantially faster than OWT3 on the ImageNet-C dataset (0.003270 vs. 0.006559). These results suggest that ReNC achieves a favorable balance between computational cost and performance.
}

\begin{table}[htpb!]
\footnotesize
\centering

\caption{Per-sample running time (in seconds) across all datasets, measured with Noise OOD samples.}
\label{tab:run-time}
\resizebox{0.65\columnwidth}{!}{   
\setlength{\tabcolsep}{2.5pt}
\begin{tabular}{cccccc}
\toprule[1pt]
Running   Time & CIFAR10-C & CIFAR100-C & ImageNet-C & ImageNet-R & VisDA-C \\
\cmidrule{1-6}
TEST & 0.000623 & 0.000626 & 0.001139 & 0.001188 & 0.001165 \\
BN & 0.000632 & 0.000658 & 0.001163 & 0.001242 & 0.001203 \\
TENT & 0.003121 & 0.003208 & 0.005682 & 0.005821 & 0.005640 \\
SHOT & 0.002406 & 0.002496 & 0.004015 & 0.003910 & 0.003733 \\
OSTTA & 0.001202 & 0.001230 & 0.002643 & 0.002656 & 0.003850 \\
EATA & 0.001726 & 0.001830 & 0.002824 & 0.002860 & 0.003195 \\
RMT & 0.002328 & 0.002382 & 0.003816 & 0.003570 & 0.003591 \\
CoTTA & 0.005638 & 0.004163 & 0.036937 & 0.009987 & 0.019757 \\
UniEnt & 0.001898 & 0.001905 & 0.003955 & 0.004019 & 0.003683 \\
OWT3 & 0.003217 & 0.003776 & 0.006559 & 0.006480 & 0.005694 \\
Ours & 0.001978 & 0.001872 & 0.003270 & 0.003493 & 0.004059\\
\bottomrule[1pt]
\end{tabular}
}
\vspace{-0.2in}
\end{table}

\newpage

\subsection{Parameter Analysis}
This section explores the model stability with respect to hyper-parameters. Given that identifying the optimal balance parameter $\lambda$ is still an unresolved issue in unsupervised learning, we adopt a parameter-tuning approach. Figure~\ref{fig:parameter analysis lambda} reports experiments with varying $\lambda \in \left\{1e-3, 1e-5, 1e-2, 5e-2, 1e-1, 5e-5\right\}$ using the $ACC_H$ metric. 
We observe that the performance remains stable across different parameter values on almost all the open-world benchmarks, with only minor variations. For the VisDA-C dataset, the proposed method also shows improved performance when $\lambda$ is larger than $5e-3$ for different OOD samples. In brief, the proposed method is stable under different settings of $\lambda$, indicating its robustness.

\begin{figure*}[htpb!]
\centering    
\subfloat[CIFAR10-C]{  
\includegraphics[width=0.19\columnwidth]{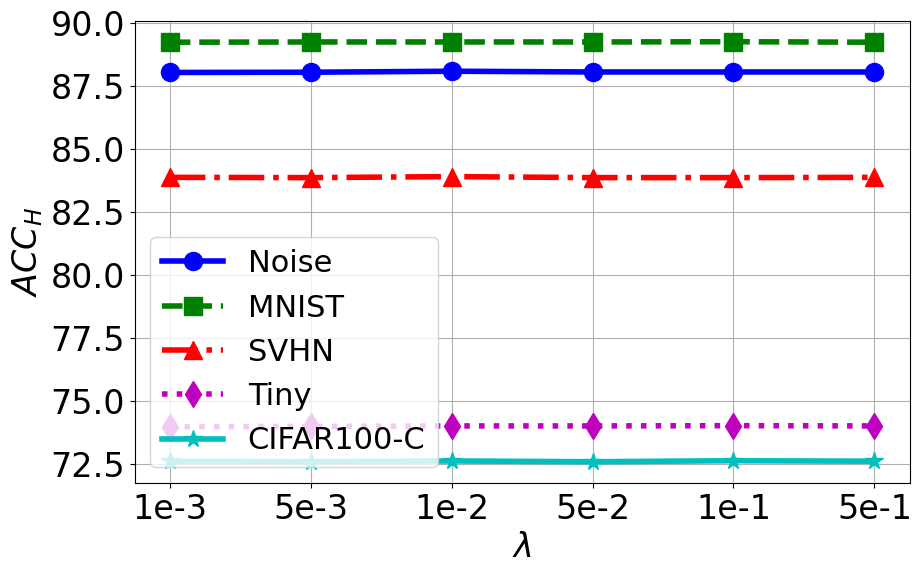}  
}  
\subfloat[CIFAR100-C]{  
\includegraphics[width=0.19\columnwidth]{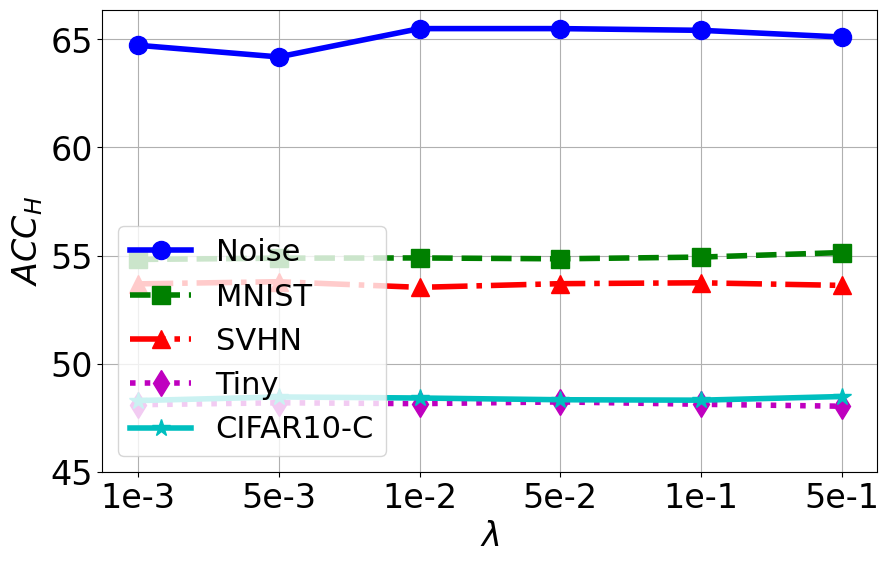}  
}
\subfloat[ImageNet-C]{      
\includegraphics[width=0.19\columnwidth]{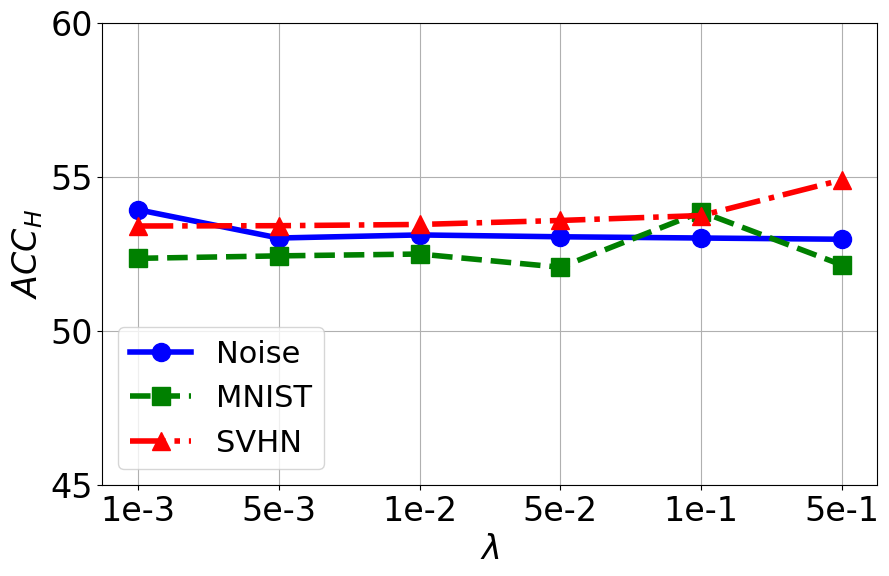}     
}
\subfloat[ImageNet-R]{  
\includegraphics[width=0.19\columnwidth]{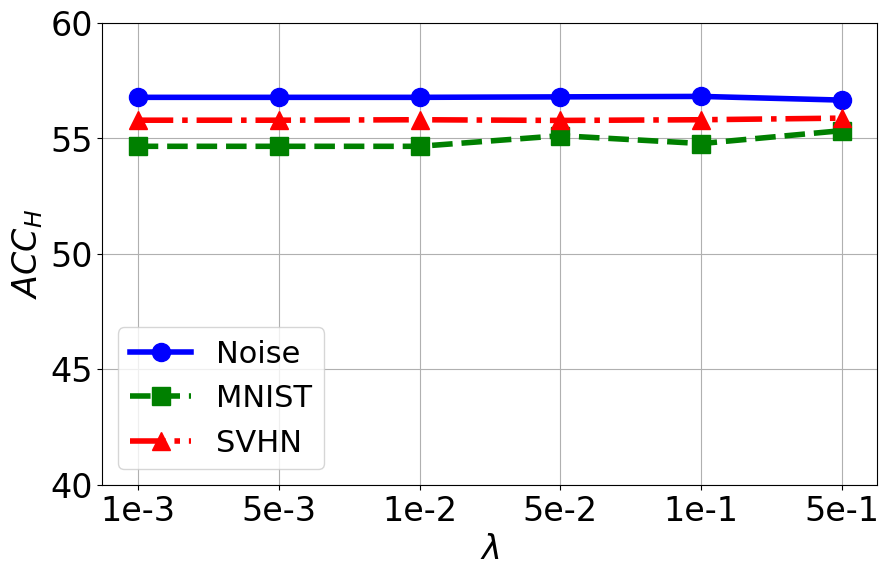}  
}  
\subfloat[VisDA-C]{  
\includegraphics[width=0.19\columnwidth]{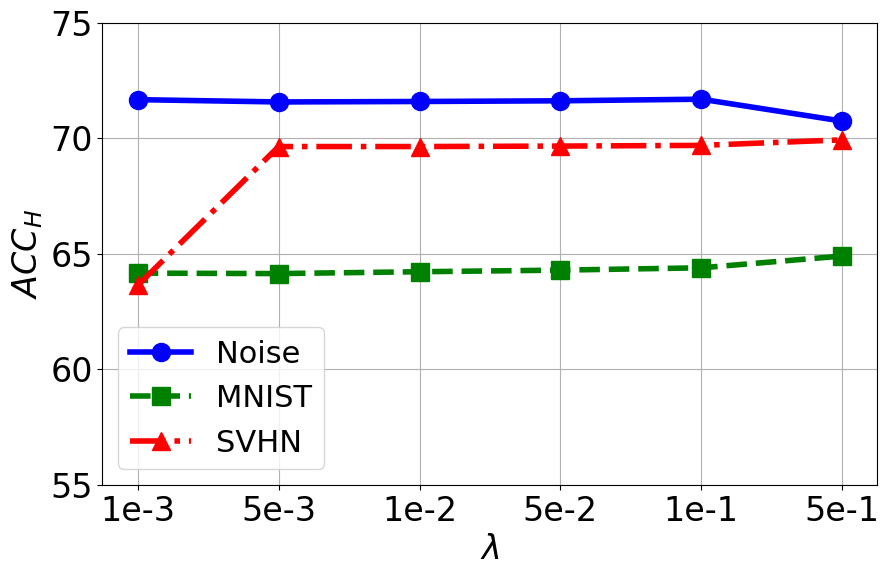}  
}  
\caption{Parameter analysis of the optimal balance parameter $\lambda$ in Eq.~\eqref{eq:obj_wo_average}.}
\vspace{-0.1cm}
\label{fig:parameter analysis lambda}
\end{figure*}

\subsection{Model Performance across Different OOD Sample Ratios}
This section validates the robustness of the proposed method against varying numbers of OOD samples. Since the ratio between ID and OOD samples can vary in real-world scenarios, we examine the impact of different OOD sample ratios. Specifically, we control the OOD sample ratio $\frac{|\mathcal{S}_{T_O}|}{|\mathcal{S}_{T_I}|}$ from 0.2 to 1.0 across all the {open-world benchmarks}, evaluating their $ACC_H$ values. The results, presented in Figure~\ref{fig:ratio}, show that the proposed method maintains consistent performance across different OOD sample ratios, demonstrating its applicability to a variety of data ratio scenarios.

\begin{figure*}[htpb!]
\centering    
\subfloat[CIFAR10-C]{  
\includegraphics[width=0.19\columnwidth]{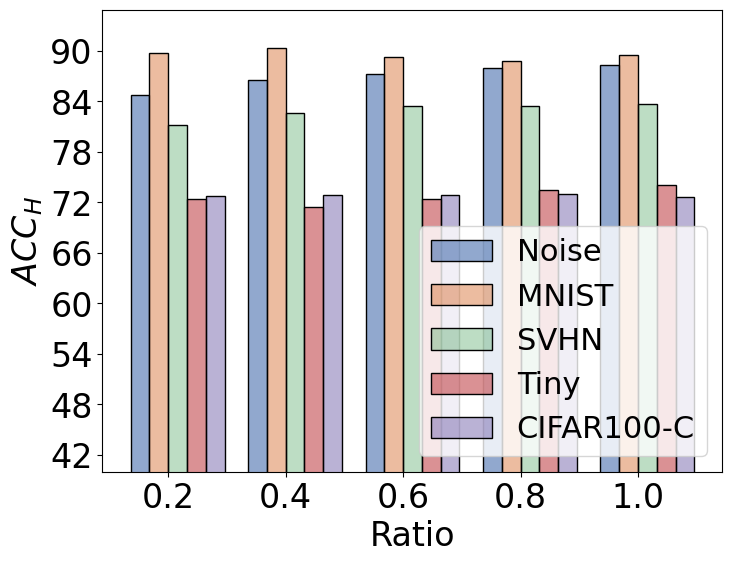}  
}  
\subfloat[CIFAR100-C]{  
\includegraphics[width=0.19\columnwidth]{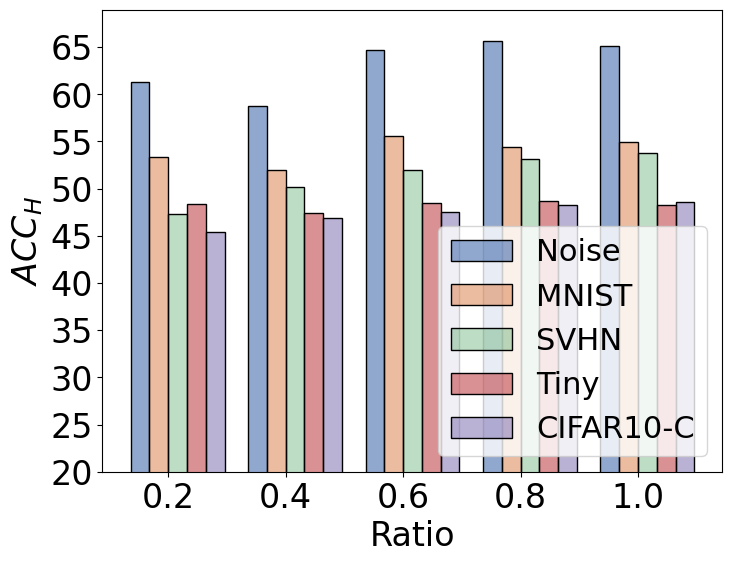}  
}
\subfloat[ImageNet-C]{      
\includegraphics[width=0.19\columnwidth]{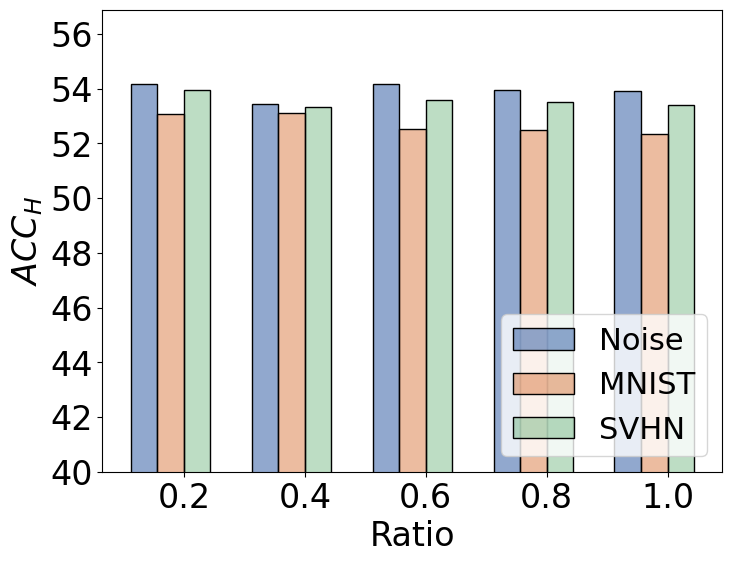}     
}
\subfloat[ImageNet-R]{  
\includegraphics[width=0.19\columnwidth]{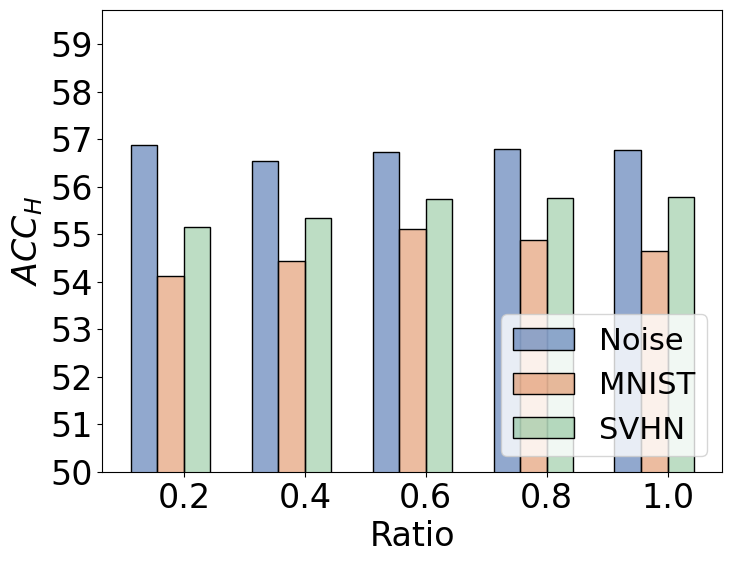}  
}  
\subfloat[VisDA-C]{  
\includegraphics[width=0.19\columnwidth]{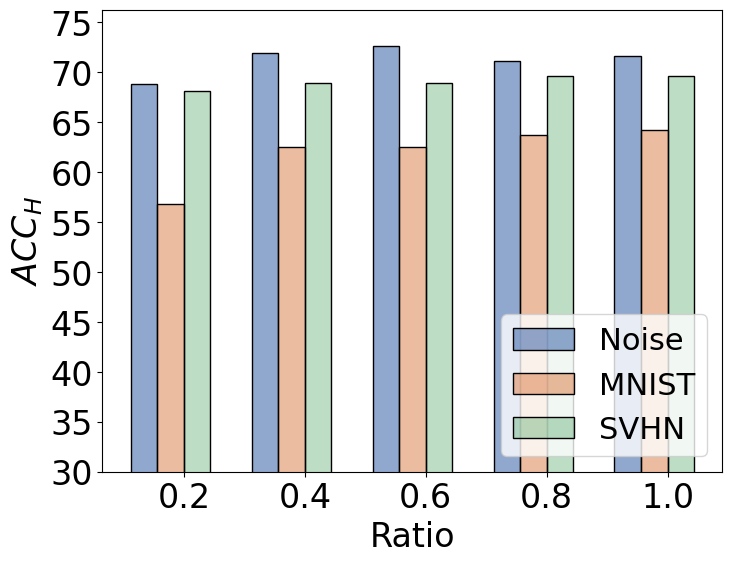}  
}  
\caption{{Open-world test-time adaptation results of ReNC at different OOD sample ratios.}}
\vspace{-0.1cm}
\label{fig:ratio}
\end{figure*}

\subsection{{Evaluation of Effectiveness on Continual OWTTA}}
{In this section, we evaluate the effectiveness of ReNC under the continual OWTTA setting, where data distribution shifts occur and OOD samples continuously change during the adaptation process. Specifically, CIFAR10-C is used as the ID dataset. For the OOD samples, we introduce the Noise, MNIST, SVHN, Tiny ImageNet, and CIFAR100-C datasets, all presented sequentially in an online manner.
The comparison results against all baseline methods are presented in Table~\ref{c_OWTTA}, using $ACC_H$ as the evaluation metric. As shown in the table, the proposed ReNC achieves the highest average performance on the continual CIFAR10-C open-world benchmark, demonstrating its strong adaptability and effectiveness in handling the continual OWTTA scenario.}

\begin{table}[htpb!]
\centering
\footnotesize
\caption{{Continual OWTTA results on CIFAR10-C with OOD samples. Best marked in bold, second best underlined.}}\label{c_OWTTA}
\centering
\setlength{\tabcolsep}{3pt}

\begin{tabular}{cccccc>{\columncolor{gray!30}}c}
 \multicolumn{7}{c}{$t\xrightarrow{\quad\quad\quad\quad\quad\quad\quad\quad\quad\quad\quad\quad\quad\quad}$}  \\
\toprule
CIFAR10-C & Noise & MNIST & SVHN & Tiny & CIFAR100-C & Average \\
 \cmidrule{1-7}
TEST & 81.78 & 75.10 & 74.85 & 68.38 & 68.07 & 73.64 \\
BN & 47.77 & 64.19 & 74.18 & 71.72 & 71.32 & 65.83 \\
TENT & 50.60 & 67.37 & 76.44 & 72.75 & 71.76 & 67.78 \\
SHOT & 50.15 & 67.74 & 77.16 & 72.97 & 71.89 & 67.98 \\
OSTTA & 21.80 & 43.70 & 46.89 & 48.22 & 47.47 & 41.62 \\
EATA & 49.76 & 64.36 & 74.09 & 71.69 & 71.34 & 66.25 \\
RMT & 44.90 & \textbf{88.10} & \underline{82.39} & \textbf{76.32} & \textbf{76.32} & 73.61 \\
CoTTA & 47.03 & 59.90 & 73.79 & 71.41 & 71.01 & 64.63 \\
UniEnt & 47.48 & 63.99 & 73.94 & 71.77 & 71.28 & 65.69 \\
OWT3 & \underline{81.93} & 75.18 & 74.94 & 68.45 & 68.08 & \underline{73.72} \\
Ours & \textbf{88.35} & \underline{86.60} & \textbf{88.62} & \underline{75.63} & \underline{72.11} & \textbf{82.26}\\
\bottomrule
\end{tabular}
\end{table}

\subsection{{Effectiveness with ViT backbone}}
{In this section, we present additional experiments using a ViT~\cite{DBLP:conf/iclr/DosovitskiyB0WZ21} backbone.
Specifically, we first pretrain a ViT model on the CIFAR10 training set as the source domain model, and then perform OWTTA experiments on the CIFAR10‑C dataset with OOD samples. Table~\ref{tab:ViT} summarizes the results in terms of $ACC_H$. As shown, ReNC consistently outperforms all baseline methods, demonstrating its robust generalization capability across different backbone architectures.}

\begin{table}[htpb!]
\centering

\caption{{OWTTA results under ViT backbone, evaluated by $ACC_H$. Best marked in bold, second best underlined.}}
\label{tab:ViT}
\begin{tabular}{cccccc}
\toprule
CIFAR10-C & Noise & MNIST & SVHN & Tiny & CIFAR100-C \\
\cmidrule{1-6}
TEST & 92.99 & 91.04 & 90.40 & 86.16 & 81.60 \\
BN & 91.81 & 90.57 & 87.89 & 84.80 & 80.34 \\
TENT & 89.15 & 92.47 & 90.89 & \underline{86.23} & \underline{82.12} \\
SHOT & 93.61 & 92.37 & 90.36 & 86.21 & 82.11 \\
OSTTA & 37.21 & 44.75 & 49.78 & 50.36 & 50.60 \\
EATA & 90.26 & 89.55 & 88.94 & 85.55 & 81.12 \\
RMT & 90.49 & 90.03 & 89.32 & 85.85 & 81.86 \\
CoTTA & 93.04 & 91.81 & 89.90 & 83.58 & 81.66 \\
UniEnt & 93.46 & 90.12 & 85.71 & 83.58 & 79.65 \\
OWT3 & \underline{93.67} & \underline{92.64} & \underline{90.94} & 86.21 & 81.92 \\
Ours & \textbf{94.88} & \textbf{93.40} & \textbf{91.61} & \textbf{87.06} & \textbf{82.72}\\
\bottomrule
\end{tabular}
\end{table}

\begin{table}[tpb!]

\caption{{Comparison of vision-language model extensions on ImageNet-R with OOD samples, evaluated by $ACC_H$. Best marked in bold, second best underlined.}}
\centering
\begin{tabular}{cccc}
\toprule
ImageNet-R & Noise & MNIST & SVHN \\
\cmidrule{1-4}
TEST & 39.63 & 38.17 & 36.61 \\
BN & 27.70 & 30.52 & 31.17 \\
TENT & 28.57 & 32.11 & 32.71 \\
SHOT & 29.28 & 31.72 & 33.29 \\
OSTTA & 35.50 & 30.93 & 35.58 \\
EATA & 27.55 & 31.07 & 31.21 \\
RMT & 16.02 & 21.70 & 27.02 \\
CoTTA & 26.35 & 28.98 & 28.92 \\
UniEnt & 18.43 & 25.26 & 26.64 \\
OWT3 & 28.07 & 31.15 & 31.75 \\
C-TPT+ & \underline{45.39} & \underline{42.40} & \underline{40.38} \\
Ours & \textbf{56.77} & \textbf{54.65} & \textbf{55.78}\\
\bottomrule
\end{tabular}
\label{tab:CLIP}
\end{table}

\subsection{{Vision Language Models in OWTTA}}
{In this section, we explore the applicability of Vision-Language Models (VLMs)~\cite{DBLP:conf/icml/RadfordKHRGASAM21} to OWTTA. While VLMs are well known for their strong zero-shot capabilities, they lack native OOD detection mechanisms and thus cannot be directly applied to the OWTTA setting. To address this limitation, we extend C-TPT~\cite{DBLP:conf/iclr/YoonYTHLY24}, a CLIP-based test-time adaptation method, by incorporating a pre-trained ResNet-50 as the OOD detection module, resulting in C-TPT+.
}

{As shown in Table~\ref{tab:CLIP}, C-TPT+ ranks second on the ImageNet-R open-world benchmark, demonstrating its strong generalization ability. This result highlights the potential of VLM-based approaches in OWTTA. 
 Notably, our ReNC is specifically designed for this task, with tailored components that support more effective learning and adaptation. As a result, it achieves the best overall performance. These findings suggest that while VLMs offer impressive zero-shot and generalization capabilities, they still require task-specific adaptation to perform reliably in specialized scenarios like OWTTA.
}

\subsection{Friedman Test for Significant Test}

 In this section, we conduct a significant test to evaluate the performance variations among the eight methods. Specifically, we utilize the Friedman test~\cite{DBLP:journals/jmlr/Demsar06} under the assumption of equal performance across all methods as the null hypothesis. Pairwise comparisons are subsequently carried out using the Nemenyi post-hoc test~\cite{nemenyi1963distribution}. The dataset for this evaluation consists of {627} instances, with each method independently assessed across nineteen datasets using three evaluation metrics. The Friedman test yields a test statistic of {$\tau_F = 18.266$}, which surpasses the critical value of {$F_{10,560} = 1.847$} at a significance level of $\alpha = 0.05$. Thus, the null hypothesis is rejected, confirming that there are statistically significant differences among the eight methods at a significance level of 0.05. 
 Further examination via the Nemenyi post-hoc test (See Figure~\ref{fig:Friedman_test}) shows that our ReNC method demonstrates a clear performance advantage over the seven baseline methods.

\begin{figure}[htpb!]
    \centering
    \includegraphics[width=0.4\linewidth]{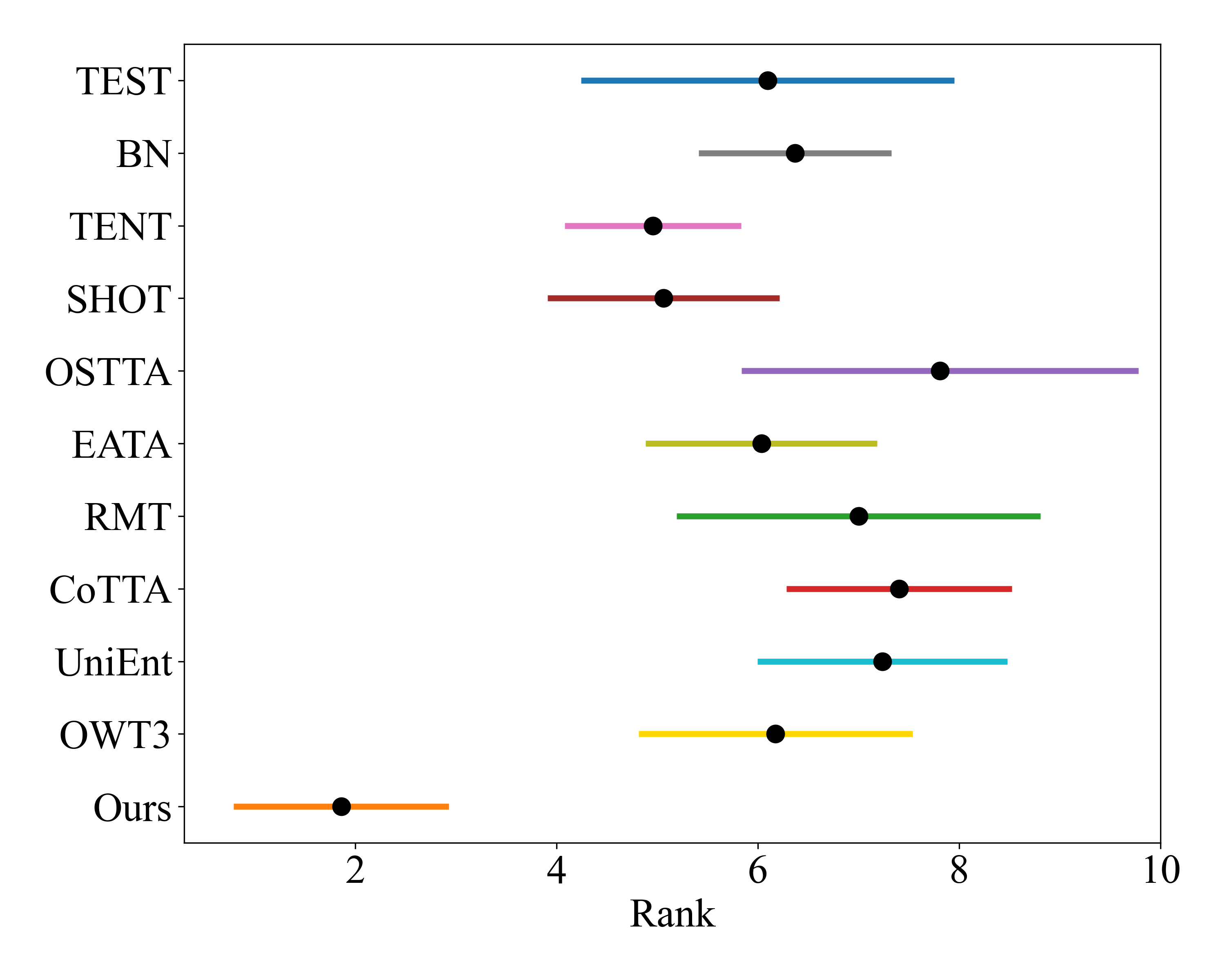}
    \caption{Nemenyi post-hoc analysis of all the baselines.}
    \label{fig:Friedman_test}
\end{figure}

\section{Conclusion}~\label{sec:conclusion}
This paper introduces Reliable Neural Collapse approximation~(ReNC), a method designed to address OWTTA reliably and effectively. We justify that pre-trained classifier weights can be viewed as prototypes within the context of neural collapse and develop a reliable update mechanism to filter out Out-Of-Distribution (OOD) samples during adaptation. Furthermore, we propose a neural collapse approximation mechanism for updating prototypes, enabling the model to gradually approach the neural collapse state in the target domain while maintaining the balanced structure. 
{Our empirical results validate the effectiveness of ReNC and suggest that NC-related properties may provide useful evidence for understanding its improved classification performance.}
{In future work, we will explore the potential of foundation models for the OWTTA.}

\bibliographystyle{IEEEtran}
\bibliography{ref}

\end{document}